\documentclass[11pt]{article}
\usepackage[utf8]{inputenc}
\usepackage[american]{babel}
\usepackage[T1]{fontenc}
\usepackage{newpxtext}
\usepackage{graphics}
\usepackage{grffile}
\usepackage{fancyhdr}
\usepackage{float}
\usepackage{amsmath}
\usepackage{amssymb}
\usepackage{amsfonts}
\usepackage[font=small,labelfont=bf]{caption}
\usepackage[export]{adjustbox}
\usepackage{mathrsfs}
\usepackage[table,xcdraw]{xcolor}
\usepackage{lipsum}
\usepackage{caption}
\usepackage{subcaption}
\usepackage{latexsym}
\usepackage{amsthm} 
\usepackage{eucal} 
\usepackage[hang]{footmisc}
\usepackage{lmodern}
\usepackage{pstricks}
\usepackage{newlfont}
\usepackage{geometry}
\usepackage{enumitem}
\usepackage{hyperref}
\renewenvironment{abstract}
 {\small
  \begin{center}
  \bfseries \Large\abstractname\vspace{-.5em}\vspace{0pt}
  \end{center}
  \list{}{
    \setlength{\leftmargin}{0.8cm}%
    \setlength{\rightmargin}{\leftmargin}%
  }%
  \item\relax}
 {\endlist}

\theoremstyle{definition}

\theoremstyle{remark}
\newtheorem{remark}{Remark}[section]
 
\providecommand{\keywords}[1]
{
  \textbf{\textit{Keywords ---}} #1
}
\usepackage{titling}
\title{System Identification Through Lipschitz Regularized Deep Neural Networks}
\author{{Elisa Negrini} \thanks{Mathematical Sciences Department, Worcester Polytechnic Institute,\newline \hspace*{1.8em} 100 Institute Road, Stratton Hall, Worcerster, MA, 01609, USA} \and Giovanna Citti \thanks{Department of Mathematics, University of Bologna \newline \hspace*{1.8em} Piazza di Porta S. Donato, 5, 40126 Bologna BO, Italy} \and Luca Capogna \protect \footnotemark[1]}
\thanksmarkseries{arabic}
\date{\vspace{-5ex}}

\begin{document}
\maketitle
\let\thefootnote\relax\footnote{\textbf{Email Addresses:} \href{mailto:enegrini@wpi.edu}{enegrini@wpi.edu} (Elisa Negrini, corresponding author),\newline \hspace*{8.0em} \href{mailto:giovanna.citti@unibo.it}{giovanna.citti@unibo.it} (Giovanna Citti),\newline \hspace*{8.0em} \href{mailto:lcapogna@wpi.edu}{lcapogna@wpi.edu} (Luca Capogna)}
\vspace{-1cm}
\begin{abstract}
In this paper we use neural networks to learn governing equations from data. Specifically we reconstruct the right-hand side of a system of ODEs $\dot{x}(t) = f(t, x(t))$ directly from observed uniformly time-sampled data using a neural network. In contrast with other neural network based approaches to this problem, we add a Lipschitz regularization term to our loss function. In the synthetic examples we observed empirically that this regularization results in a smoother approximating function and better generalization properties when compared with non-regularized models, both on trajectory and non-trajectory data, especially in presence of noise. In contrast with sparse regression approaches, since neural networks are universal approximators, we don't need any prior knowledge on the ODE system. Since the model is applied component wise, it can handle systems of any dimension, making it usable for real-world data.
\end{abstract}
\section{Introduction}
Dynamical system models are widely used to study, explain and predict behaviour in multiple application areas such as Newton's laws of mechanics, economic and financial systems, biology, medicine, social systems and so on (see for instance \cite{billings2013nonlinear} for more examples of applications). Governing laws and equations have traditionally been derived from expert knowledge and first principles, however in recent years the large amount of data available resulted in a growing interest in data-driven models and approaches for automated model discovery.\\ System identification deals specifically with the problem of building mathematical models and approximating governing equations using only observed data from the system. Since the theory of time-invariant linear system has been widely studied over the past decades, many methods for linear system identification have been developed (see for instance \cite{nelles2013nonlinear}, \cite{ljung2010perspectives}). In this work we extend this study to time dependent nonlinear systems.\\ Some frequently used approaches for data-driven discovery of nonlinear differential equations are sparse regression, Gaussian processes and neural networks. Sparse regression approaches are based on a user-determined library of candidate terms from which the most important ones are selected using sparse regression (see for instance \cite{schaeffer2013sparse}, \cite{rudy2017data}, \cite{schaeffer2017learning}). Identification using Gaussian Processes places a Gaussian prior on the unknown coefficients of the differential equation and infers them via maximum likelihood estimation (see for instance \cite{raissi2017machine}, \cite{raissi2018hidden}, \cite{raissi2018numerical}). Being universal approximators, neural networks have been widely used in nonlinear system identification: depending on the architecture and on the properties of the loss function, they can be used as sparse regression models, they can act as priors on unknown coefficients or completely determine an unknown differential operator. Many kind of architectures have been used for system identification, among which multi-layer feed forward networks (see for instance \cite{narendra1992neural}, \cite{kuschewski1993application}, \cite{berg2017neural}, \cite{raissi2018deep}, \cite{berg2019data}) and recurrent networks and its variants which have been used in dynamic identification of nonlinear systems because of their ability to retain information in time across layers (see for instance \cite{wang2006fully}, \cite{dinh2010dynamic}, \cite{ogunmolu2016nonlinear}, \cite{qin2019data}).
\medskip\\
In this work we investigate the problem of approximating unknown governing equations, i.e. the right-hand side $f$ of the system of ODEs:
\begin{equation} \label{ODE}
  \dot{x}(t) = f(t, x(t))  
\end{equation}
directly from observed trajectories using a deep neural network $N$. The main contribution of this paper is that we improve generalization and recovery of the governing equation by adding a Lipschitz regularization term in our loss function. This term forces the Lipschitz constant of the network $N$ to be small, improving the smoothness of the approximating function even in presence of noise in the data. The choice of this specific kind of regularization is inspired by Oberman and Calder's work in \cite{oberman2018lipschitz} where they prove that Lipschitz regularized networks converge and generalize. We empirically show that the test error on trajectory data as well as the recovery error on non-trajectory data improves when using this kind of regularization.  Other works, such as \cite{xu2012robustness},  \cite{gouk2018regularisation} and \cite{pauli2020training}  also make use of a Lipschitz constraint to improve generalization and robustness; however, they differ from our work since they do not apply such regularization to system identification and they use a different approaches to estimate and bound the Lipschitz constant of the network.\\ The use of neural networks for system identification has multiple advantages: since neural networks are universal approximators, we do not need to commit to a particular dictionary of basis functions, nor need any prior information about the order or about the analytical form of the differential equation as in \cite{schaeffer2013sparse}, \cite{rudy2017data}, \cite{schaeffer2017learning}, \cite{sahoo2018learning}, \cite{hasan2020learning}; this allows to recover very general differential equations.
%but it comes at the cost of losing interpretability of the learned dynamic.
Another advantage of our approach is that, thanks to the Lipschitz regularization term, our network is able to generalize better than non-regularized networks for unseen data (both for trajectory and non-trajectory data) and it is robust to noise. This is especially an advantage over works that use finite differences and polynomial approximation to extract governing equations from data (for instance \cite{brunton2016discovering}, \cite{rudy2017data}) which perform poorly in presence of noise. Finally, our model is defined componentwise so it can be applied to system of equations of any dimension, thus making it a valuable approach when dealing with high dimensional real-world data.\\
Specifically, we consider the system of ODEs (\ref{ODE}),
where $x(t) \in \mathbb{R}^d$ is the state vector of a $d$-dimensional dynamical system at time $t \in I \subset \mathbb{R}$, $\dot{x}(t) \in \mathbb{R}^d$ is the first order time derivative of $x(t)$ and $f: \mathbb{R}^{1+d}\rightarrow \mathbb{R}^d$ is a
vector-valued function right-hand side of the differential equation. We approximate the unknown function $f$ with a neural network $N$. Our observables are the trajectories corresponding to the state vectors $x(t)$ sampled at discrete uniformly-spaced time steps for different initial conditions. We note that, since our
neural network produces a Lipschitz regular approximant, our approach is most useful when the function $f$ is Lipschitz continuous (corresponding to a deterministic governing law), however we are not assuming anything about its analytical form.\\The loss function we minimize to learn $f$ is:
\begin{equation}\label{Loss_lip}
    L(\theta) =\frac{1}{k}\sum_{i=1}^k l( N( X_i, \theta), Y_i ) + \alpha \text{Lip}(N(\cdot, \theta)),
\end{equation}
where $\theta$ are the network parameters, $\alpha >0$, $(X_i, Y_i)$ are the training data and Lip$(N(\cdot, \theta))$ is the Lipschitz constant of the network as a function of the inputs. The targets $Y_i$ are approximations of $\dot{x}(t)$, the left-hand side of (\ref{ODE}) obtained by difference quotients for non-noisy data or polynomial interpolation in presence of noise. We stress that accurate interpolation of measurement data is necessary to obtain reliable derivative approximation which are used as target data. The network resulting from this minimization will reach a trade off between fitting the data (by minimizing the first term in (\ref{Loss_lip})) and having a small Lipschitz constant (by minimizing the second term in (\ref{Loss_lip})). This second characteristic is the one that makes our model robust to noise and able to generalize well for unseen data.
We evaluate the generalization capabilities of our model by means of the well known generalization gap (see \cite{abu2012learning}).
\medskip\\
The paper is organized as follows: in Section 2 we describe the data structure, pre-proccesing and selection of training and testing data; in Section 3 we describe the neural network we used, the loss function and we define the generalization gap; in Section 4 we propose numerical examples and discuss the results; in Section 5 we show an example of recovery of the right-hand side function on non-trajectory data; in the conclusion Section we summarize our results and describe possible future directions of research.
\newpage
\section{Data}
In this section we describe the data structure, preprocessing and selection of training and testing data.
\medskip\\
As explained above, our goal is to recover the right-hand side $f$ of the system of ordinary differential equations (\ref{ODE}) from discrete observations of the state vector $x(t) \in \mathbb{R}^d$.\\ %through an approximation with a Lipschitz regularized deep neural network $N$.
Given equally spaced time points $t_1,\dots, t_M$ and initial conditions $x_1(0), \dots, x_K(0) \in \mathbb{R}^d$, define $x_i(t_j) \in \mathbb{R}^d, \quad i = 1,\dots,K, \quad j = 1,\dots, M,\;$ to be an observation of the state vector $x(t)$ at time $t_j$ for initial condition $x_i(0)$. The recovery of $f$ is obtained through an approximation with a Lipschitz regularized deep neural network $N$;\; the targets $\dot{x}_i(t_j) \in \mathbb{R}^d$ are the first derivative approximation of $x(t)$ at time $t_j$ for initial condition $x_i(0)$ and are generated using the state vector $x(t)$.
\medskip\\
Specifically, the network data used in our model are couples $(X_h, Y_h), \; h = j + (i-1)M = 1, \dots, KM$, where $X_h$ is the  input and $Y_h$ is the target and $X_h$, $Y_h$ are defined as follows:
\begin{align*}
&X_h = (t_j, \; x_i^1(t_j), \; x_i^2(t_j),\; \dots \; x_i^d(t_j)) \in \mathbb{R}^{1+d},\\
&Y_h = \dot{X_h} = (\dot{x}_i^1(t_j), \; \dot{x}_i^2(t_j),\; \dots \; \dot{x}_i^d(t_j)) \in \mathbb{R}^{d}.
\end{align*}
The data is separated into training and testing sets made respectively of 80\% and 20\% of the data.\\
\begin{figure}[H]
\centering
\begin{subfigure}{.5\textwidth}
\captionsetup{width=.6\linewidth}
  \centering
  \includegraphics[width=.7\linewidth]{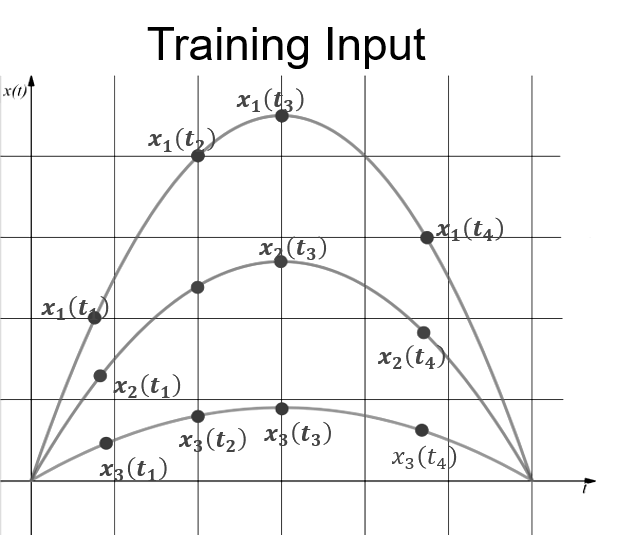}
  \caption{Observations of $x(t)$ for initial conditions $x_1(0),\, x_2(0),\, x_3(0)$ and time points $t_1,\dots, t_4$\vspace{18pt}}
  \label{fig:sub1}
\end{subfigure}%
\begin{subfigure}{.5\textwidth}
\captionsetup{width=.6\linewidth}
  \centering
  \includegraphics[width=.7\linewidth]{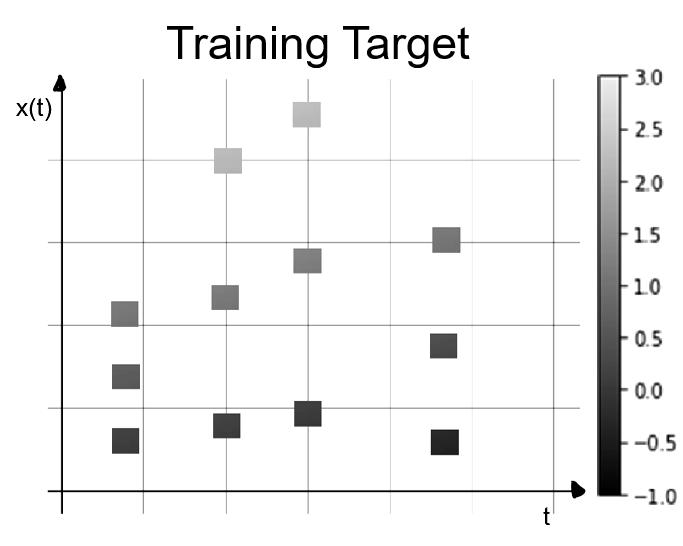}
  \caption{Approximations of $\dot{x}(t)$ for initial conditions $x_1(0),\, x_2(0),\, x_3(0)$ and time points $t_1,\dots, t_4$, the color represents the value of $\dot{x}(t)$}
  \label{fig:sub2}
\end{subfigure}
\caption{Example of Training Data}
\label{fig:test}
\end{figure}
\begin{figure}[H]
\centering
\begin{subfigure}{.5\textwidth}
\captionsetup{width=.8\linewidth}
  \centering
  \includegraphics[width=0.9\linewidth, height=0.9\linewidth]{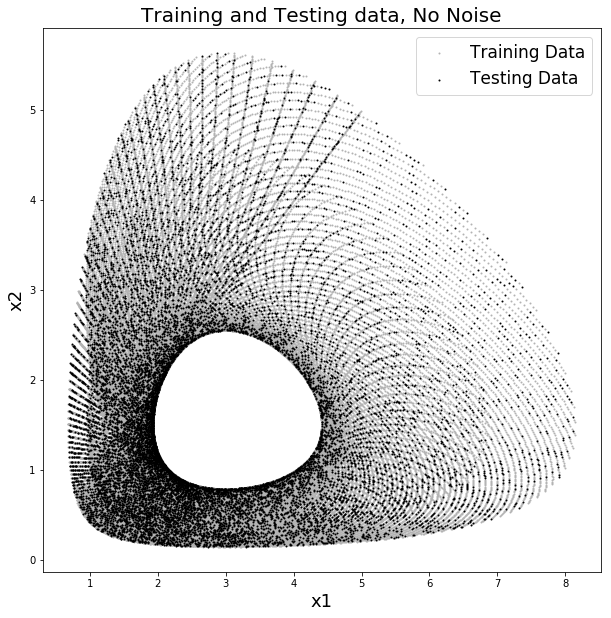}
  \caption{Example of Distribution of Training and Testing Data, No Noise. Gray points are training data, black points are testing data.}
  \label{fig:train_test0N}
\end{subfigure}%
% \begin{subfigure}{.33\textwidth}
% \captionsetup{width=.8\linewidth}
%   \centering
%   \includegraphics[width=0.95\linewidth,, height=0.95\linewidth]{train_test_1N.png}
%   \caption{Example of Distribution of Training and Testing Data, 1\% Noise}
%   \label{fig:train_test1N}
% \end{subfigure}
\begin{subfigure}{.5\textwidth}
\captionsetup{width=.8\linewidth}
  \centering
  \includegraphics[width=0.9\linewidth,, height=0.9\linewidth]{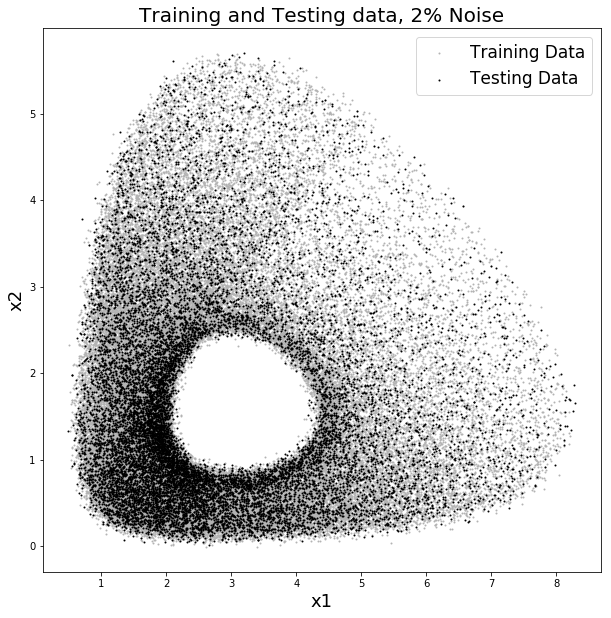}
  \caption{Example of Distribution of Training and Testing Data, 2\% Noise. Gray points are training data, black points are testing data.}
  \label{fig:train_test2N}
\end{subfigure}
\caption{Distribution of Training and Testing Data for Different Amounts of Noise. Gray points are training data, black points are testing data.}
\label{fig:train_test}
\end{figure}
In the numerical examples we use synthetic data generated in Python:  using the function \texttt{odeint} from the \texttt{scipy} package in Python, we solve $\dot{x}(t) = f(t, x(t))$;\; this provides us with approximations of the state vector $x(t)$ for initial conditions  $x_1(0), \dots, x_K(0) \in \mathbb{R}^d$ at time steps $t_1,\dots, t_M$. We perform the experiments in the case of noiseless data, and data with $1\%$ and $2\%$ of noise. To generate noisy data, we proceed as follows: for each component $x^k(t)$ of the solution $x(t)$ we compute its mean range $M_k$ across trajectories as $$M_k =\frac{1}{K}\left(\sum_{i=1}^K |\max_{j=1,\dots,M} x_i^k(t_j) - \min_{j=1,\dots,M} x_i^k(t_j)|\right).$$ 
Then, the $1\%$ noisy version of $x_i^k(t_j)$ is given by $$\hat{x}_i^k(t_j) = x_i^k(t_j) + n_{ij}M_k, $$ where $n_{ij}$ is a sample from a normal distribution $\mathscr{N}(0,0.01)$ with mean 0 and variance 0.01. In a similar way we add $2\%$ of noise to the data.
\medskip\\
The next step is to generate approximations of the first order time derivative of $x(t)$ by approximating each component $\dot{x}^k(t)$ of $\dot{x}(t)$ using difference quotients. The difference quotient approximation of the derivatives can be highly inaccurate at the boundaries of the time interval as well as not reliable for a large time step $\Delta t$ or in presence of noise. To deal with these issues, we first preprocess our data by extending it on the left and on the right of the time interval using odd extension to improve accuracy of the derivative approximation at the boundaries. Then, for noisy data we interpolate componentwise the noisy state vector $\hat{x}_i(t_j)$ using cubic splines and we compute derivative approximations using the smoothed data points. Note that the input data for our network in the noisy case is still the original noisy state vector $\hat{x}_i(t_j)$: the smoothed spline version of the state vector is only used to generate reliable target data.\\This preprocessing has multiple advantages: the spline interpolation of the state vector gives a smooth approximation of the solution vector which greatly improves the quality of the derivative approximation obtained with difference quotients, especially in presence of noise in the data; the odd extension at the boundaries preserves the time derivative at the first and last time step and allows us to use difference quotients to approximate the derivatives of the state vector at boundary times. Many other options for data preprocessing are present in literature (see for example \cite{garcia2015data}, \cite{kotsiantis2006data} for a review), among which neural networks, which were used in this way in \cite{berg2019data}. However, spline interpolation is faster than neural network approximation and, since the sampling of the state vector is assumed to be uniform in time, it gives a reliable smooth approximation of the solution. We also want to mention that, while spline interpolation works very well on uniformly sampled data, it does not work well if the data is not well sampled. This is the reason why we successfully used splines on the state vector data, but could not solve the original reconstruction problem with splines since the sampling in space is not uniform. We leave the interesting problem of finding a better preprocessing method using neural networks for future work. We also leave for a future work the application of our method to real data for which we expect the preprocessing and denoising of the data to be a significant part of the work as explained in \cite{berg2019data}.
\medskip\\
To evaluate the performance of our network, we use the Mean Squared Error (MSE) on test data. We note that in a real-world problem the test error is the only information accessible to evaluate the performance on the model and this is why we are mostly interested in this value even if presenting only synthetic examples. However, since we used synthetic data, we have access to the real right-hand side function $f$; in Section 5 we compare the output of the network with the true function $f(t,x)$ on arbitrary couples $(t,x)$ in the domain of $f$. We will see that that the recovery error is, not surprisingly, larger than the test error. This is due to the fact that, because of its nature as position vectors along a trajectory, the data is uniformly sampled in the time variable $t$, but in general, it is not uniformly sampled in the space variable $x$, so that the domain of $f$ may not be well covered by the data. We will see that despite being larger than the test error, the best recovery error is always attained by a regularized network, showing once again that Lipschitz regularization improves generalization.
% \begin{figure}[H]
% \captionsetup{width=.5\linewidth}
% \centering
% \includegraphics[width=.4\linewidth]{Net image.png}
% \caption{Output of the network computed on $(t,x)$ couples in the domain of $f$.}
% \label{fig:target}
% \end{figure}
\section{The Model}
In this section we describe the neural network used in the experiments and the loss function.
\smallskip\\
In the numerical experiments we use 
a feed forward neural network with $L$ hidden layers and Leaky ReLU activation function to approximate the right-hand side of an ordinary differential equation. We apply the network to each training input $X_h$ and we aim to find the best network parameters to match the corresponding $Y_h$.
\smallskip\\
For $ i = 1, \dots, L$ define the weight matrices $W_i \in \mathbb{R}^{\;n_i \times n_{i-1}}$ and bias vectors $b_i \in \mathbb{R}^{n_i}$ where $n_i \in \mathbb{N}, n_0 = 1+d, n_L = d$. Let $\theta = \{W,b\}$ be the model parameters.\\
As activation function, we use a Leaky Rectified Linear Unit (LReLU) with parameter $\varepsilon = 0.01$, defined as:
\begin{align*}
    \sigma(x) = \text{LReLU}(x) = \begin{cases}
    \varepsilon x &\text{if } x<0;\\
    x &\text{if } x\geq0.
    \end{cases}
\end{align*}
For an input $X_h \in \mathbb{R}^{1+d}$ and parameters $\theta$ we have: 
$$N(X_h, \theta) = (\dots (\sigma ( \sigma ( X_h W^T_1 +b_1)W_2^T +b_2) \dots)W_L^T + b_L \; \in \mathbb{R}^{d}.$$
\begin{remark}
For ease of notation, in the rest of the paper we will drop the explicit dependence of the network $N$ from the parameters $\theta$ and we will only write $N(X_h)$ instead of $N(X_h, \theta)$.
\end{remark}
% \begin{figure}[h]
%     \centering
%     \includegraphics[width=0.7\linewidth]{FFN_bw.png}
%     \caption{Feed Forward Network with $L$ layers}
%     \label{fig:net}
% \end{figure}
\begin{remark}
The choice of LReLU as activation function is due to the fact that Multilayer Feed Forward Networks with LReLU activation functions are dense in the set of continuous functions on $\mathbb{R}^{n_0}$. This is an application of Theorem 1 in \cite{leshno1993multilayer} which states that if the activation function is $L^{\infty}_{loc}(\mathbb{R})$, it is not an algebraic polynomial almost everywhere and the closure of the set of its discontinuity points has measure zero, then the class of functions implemented by a multilayer feed forward network is dense in $C(\mathbb{R}^{n_0})$.\\ Since our
neural network produces a Lipschitz regular approximant, our approach is most useful when the function $f$ is Lipschitz continuous. In this case, in fact, since neural networks with LReLU activation function are dense in the set of continuous functions on $\mathbb{R}^{n_0}$, it is possible in theory, given a large enough number of layers and nodes, to approximate the function $f$ within any prescribed error.
%In this paper we assume existence and uniqueness (given the initial condition) of the solution of $\dot{x}(t) = f(t, x(t))$, thus we are implicitly assuming that the function $f$ which we are trying to approximate using a neural network is Lipschitz continuous.
% Note, however, that the theorems proved in \cite{leshno1993multilayer} only ensure the existence of a LReLU network which, for any given $\varepsilon >0$, approximates $f$ within $\varepsilon$; since we use gradient descent methods to minimize the loss function, we may not be able to obtain the specific parameters (which exist by the theorems in \cite{leshno1993multilayer}) for which our network approximates $f$ with an error of almost $\varepsilon$.
\end{remark}
\subsection{The Loss Function}
In our experiments the loss function is the Mean Squared Error (MSE) with a Lipschitz regularization on the network. In contrast with the most common choices of regularization terms found in Machine Learning literature, we don't impose an explicit regularization on the network parameters, but we impose a Lipschitz regularization on the statistical geometric mapping properties of the network. Of course, since the minimization of the loss function is done with respect to the parameters, the Lipschitz regularization term will result in an implicit constraint on the network parameters.\\ The loss function we use is defined as follows:
$$L(\theta) = \frac{1}{KM}\sum_{h=1}^{KM} \| Y_h - N(X_h) \|^2_2 + \alpha \text{Lip}(N),$$
where $\| \cdot \|_2$ is the $L^2$ norm, $\alpha >0$ is a regularization parameter and $\text{Lip}(N)$ is the Lipschitz constant of the network $N$. The predicted approximation of the function $f(t,x)$ is given by the network $N$ corresponding to $\underset{\theta}{\mathrm{argmin}}\, L(\theta)$.\\
The network $N$ implements a function $N:\mathbb{R}^{1 + d} \rightarrow \mathbb{R}^d$ and its Lipschitz constant
$\text{Lip}(N)$ can be computed as:
$$\text{Lip}(N) = \sup_{x \in \mathbb{R}^{d + 1}} \sup_{y \in \mathbb{R}^{d + 1}}\frac{\|N(x) -N(y)\|_2}{\|x -y\|_2}.$$
In practice, we approximate the Lipschitz constant of the network using a similar approach to the one presented in \cite{calliess2015bayesian}: we randomly select a finite set $S$ of points on the training trajectories; then, we estimate the Lipschitz constant of the network as 
$$\text{Lip}(N) = \sup_{x \in S} \sup_{y \in S}\frac{\|N(x) -N(y)\|_2}{\|x -y\|_2}.$$
Note that, as empirically shown in \cite{calliess2015bayesian}, the larger the cardinality of the set $S$, the better the approximation of the Lipschitz constant.\\
We explicitly note that controlling the Lipschitz constant of the network $N$ yields control on the smoothness and rate of change of the approximating function.

\subsection{Generalization Gap}
As explained before, the goal of this paper is to compare the performance of Lipschitz regularized networks and non-regularized networks when approximating unknown governing equations from observed data. To do this, we perform multiple numerical experiments and for a fixed training error we compare the test error and generalization gap for multiple choices of the Lipschitz regularization parameter. To be more precise we define $\rho$ to be the true data distribution, $\mathscr{D}_k$ the training data distribution, discrete approximation of $\rho$ which converges to $\rho$ as the number of data points $k$ tends to infinity, and $\mathscr{D}_{test}$ to be the discrete distribution of test data. We write $X \sim \rho$ to indicate that the random variable $X$ has distribution $\rho$.
\medskip\\
Generalization gap in machine learning refers to the difference:
$$\mathbb{E}_{X \sim \rho}[\| N_k(X) - Y(X) \|^2_2] - \mathbb{E}_{X \sim \mathscr{D}_k}[\| N_k(X) - Y(X) \|^2_2],$$
where $N_k$ denotes the function learned when using $k$ data points after minimizing the loss function $$L(\theta) =\frac{1}{k}\sum_{i=1}^k \| N_k(X) - Y(X) \|^2_2 + \alpha \text{Lip}(N)$$ on the training data $\mathscr{D}_k$. The quantity $\mathbb{E}_{X \sim \mathscr{D}_k}[\| N_k(X) - Y(X) \|^2_2]$ is known since it can be evaluated on the training data $\mathscr{D}_k$ once the optimal $N_k$ has been found.\\ On the other hand, the quantity $\mathbb{E}_{X \sim \rho}[\| N_k(X) - Y(X) \|^2_2]$ is unknown since we do not have access to the true data distribution $\rho$. This quantity, however, can be estimated by using a test set of data, $\mathscr{D}_{\text{test}}$, which is a data set that was not used in the training process.\\ The network $N_k$ which minimizes the training error is evaluated on the test set an the value of  $\mathbb{E}_{X \sim \mathscr{D}_{\text{test}}}[\| N_k(X) - Y(X) \|^2_2]$, is taken as an estimate of $\mathbb{E}_{X \sim \rho}[\| N_k(X) - Y(X) \|^2_2]$. This estimate is only accurate if the distribution $\mathscr{D}_{\text{test}}$ is a faithful discrete representation of the true data density $\rho$, that is, if the discrete distribution $\mathscr{D}_{\text{test}}$ converges, as the number of test data goes to infinity, to the true distribution $\rho$.\\ The Hoeffding inequality (see \cite{abu2012learning}, section 1.3) gives a bound which depends on the number of test data on the approximation of $\mathbb{E}_{X \sim \rho}[\| N_k(X) - Y(X) \|^2_2]$ obtained using $\mathbb{E}_{X \sim \mathscr{D}_{\text{test}}}[\| N_k(X) - Y(X) \|^2_2]$: if $m$ is the number of test data, given any $\varepsilon >0$ the Hoeffding inequality states that:
$$ \mathbb{P}(|\mathbb{E}_{X \sim \rho}[\| N_k(X) - Y(X) \|^2_2] - \mathbb{E}_{X \sim \mathscr{D}_{\text{test}}}[\| N_k(X) - Y(X) \|^2_2]|>\varepsilon)\leq 2e^{-2 \varepsilon^2 m}.$$
So that if we have $m = 1000$ data points in the test set, then $\mathbb{E}_{X \sim \rho}[\| N_k(X) - Y(X) \|^2_2]$ will be within $5\%$ of $\mathbb{E}_{X \sim \rho}[\| N_k(X) - Y(X) \|^2_2]$ with $98\%$ of probability. This inequality justifies the use of test error in our numerical examples as an estimate of $\mathbb{E}_{X \sim \rho}[\| N_k(X) - Y(X) \|^2_2]$.
\section{Numerical Examples}
In this section we present numerical examples. We propose four examples of recovery of ordinary differential equations: a one dimensional autonomous ODE, a one dimensional time dependent ODE, the Lotka Volterra system and the pendulum equation. The results shown are representative of a larger testing activity, in which several different types of right-hand sides $f(x,t)$ have been used, leading to comparable experimental results.\\For each example we summarize the Training and Testing MSE and the Genralization Gap across all choices of the regularization parameter both in case of noiseless and noisy data. In each case we select the best regularization parameter as the one that attains the best test accuracy. To compare the best regularized model with the non-regularized one we plot graphs of the true and recovered function as well as graphs of absolute test error when using non-regularized and the best regularized models. In each example we will describe the specific choices of time interval, time step and initial conditions as well as the hyperparameters, selected by cross validation, of our network. We also notice that the one-dimensional examples are the basis for the multi-dimensional ones since the recovery in higher dimension is performed componentwise.

\subsection{Autonomous 1D ODE}
\begin{figure}[H]
\centering
\begin{subfigure}{.33\textwidth}
\captionsetup{width=.8\linewidth}
  \centering
  \includegraphics[width=1\linewidth, height=0.915\linewidth]{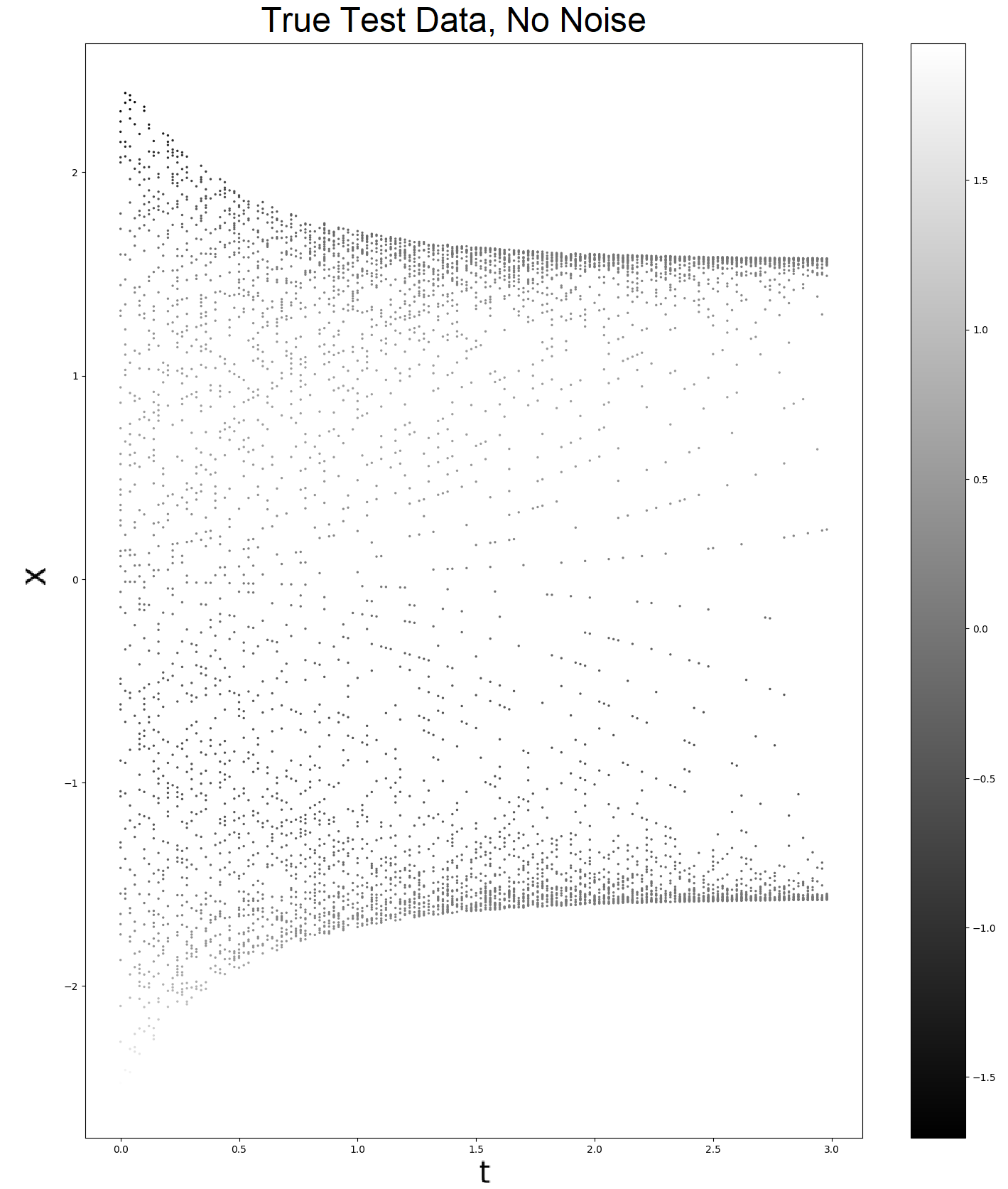}
  \caption{True test data.\vspace{24pt} }
  \label{fig:Truexcosx}
\end{subfigure}%
\begin{subfigure}{.33\textwidth}
\captionsetup{width=.8\linewidth}
  \centering
  \includegraphics[width=0.95\linewidth,, height=0.95\linewidth]{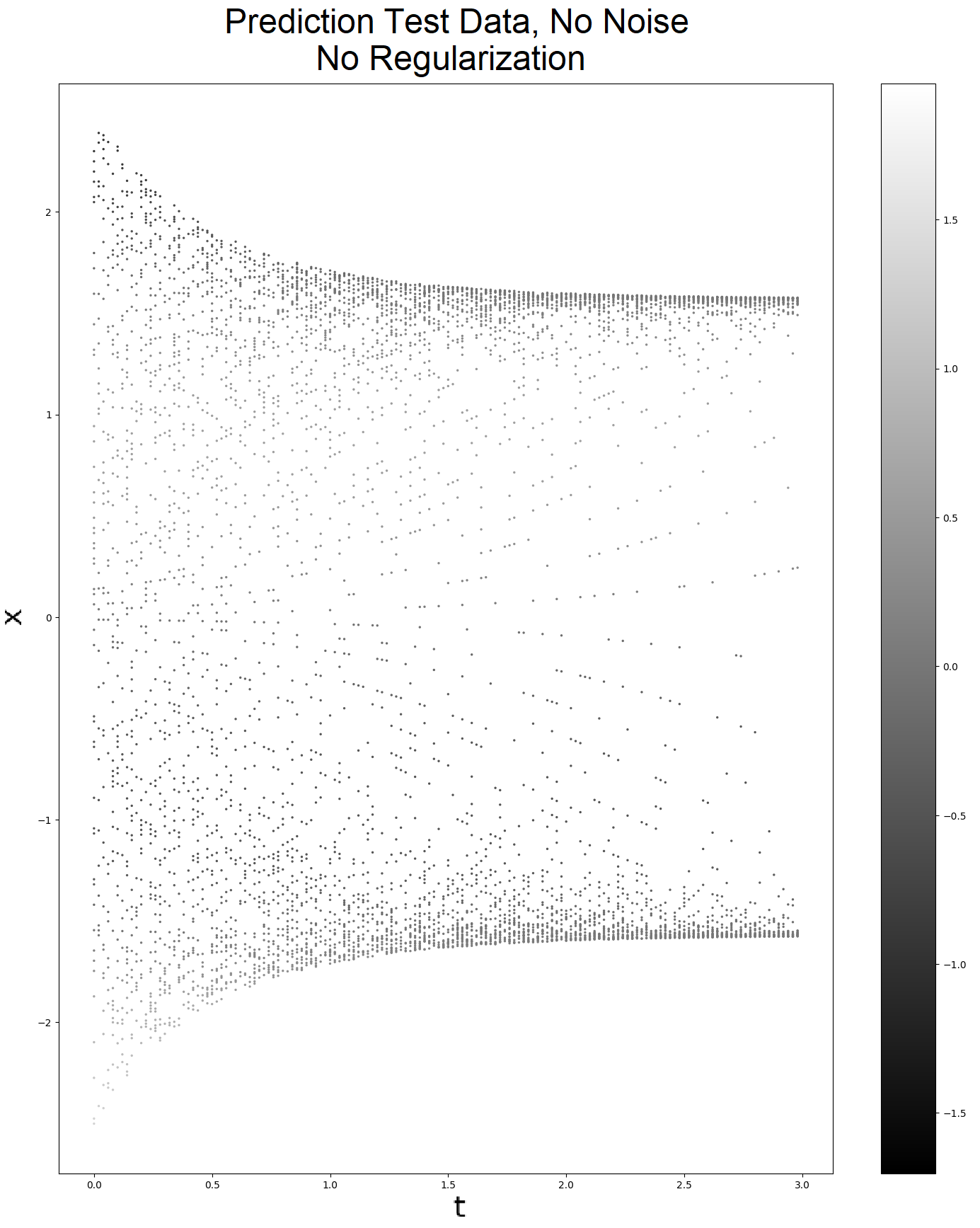}
  \caption{Test prediction, non-regularized network. \vspace{12pt}}
  \label{fig:pred0NcosxNR}
\end{subfigure}
\begin{subfigure}{.33\textwidth}
\captionsetup{width=.8\linewidth}
  \centering
  \includegraphics[width=0.95\linewidth,, height=0.95\linewidth]{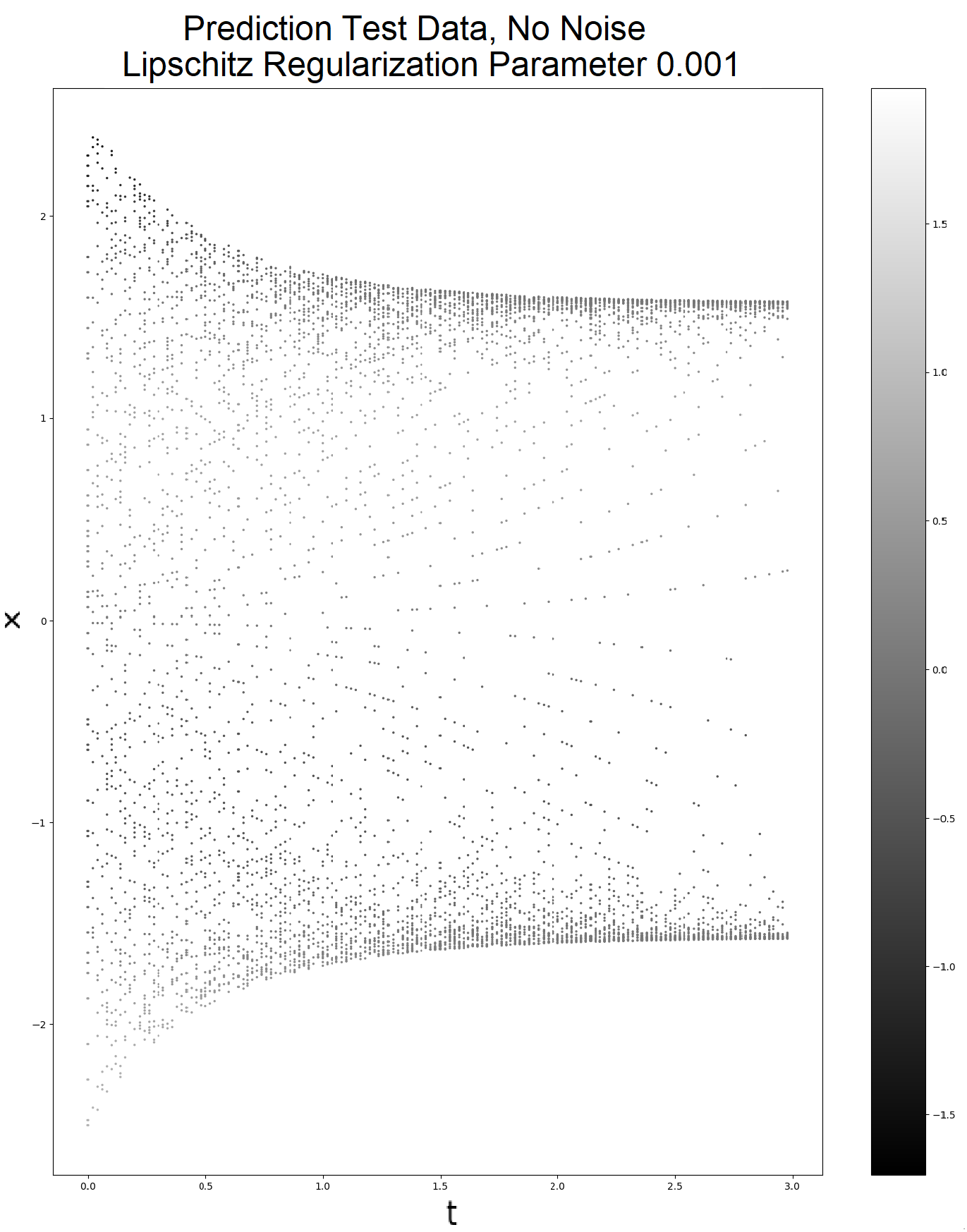}
  \caption{Test prediction, Lipschitz regularized network with parameter 0.001.}
  \label{fig:pred0NcosxLR}
\end{subfigure}
\caption{Prediction comparison on test data, noiseless data.}
\label{fig:predxcosx}
\end{figure}

\begin{figure}[H]
\centering
\begin{subfigure}{.5\textwidth}
\captionsetup{width=.8\linewidth}
  \centering
  \includegraphics[width=0.85\linewidth,height=0.85\linewidth,valign=t]{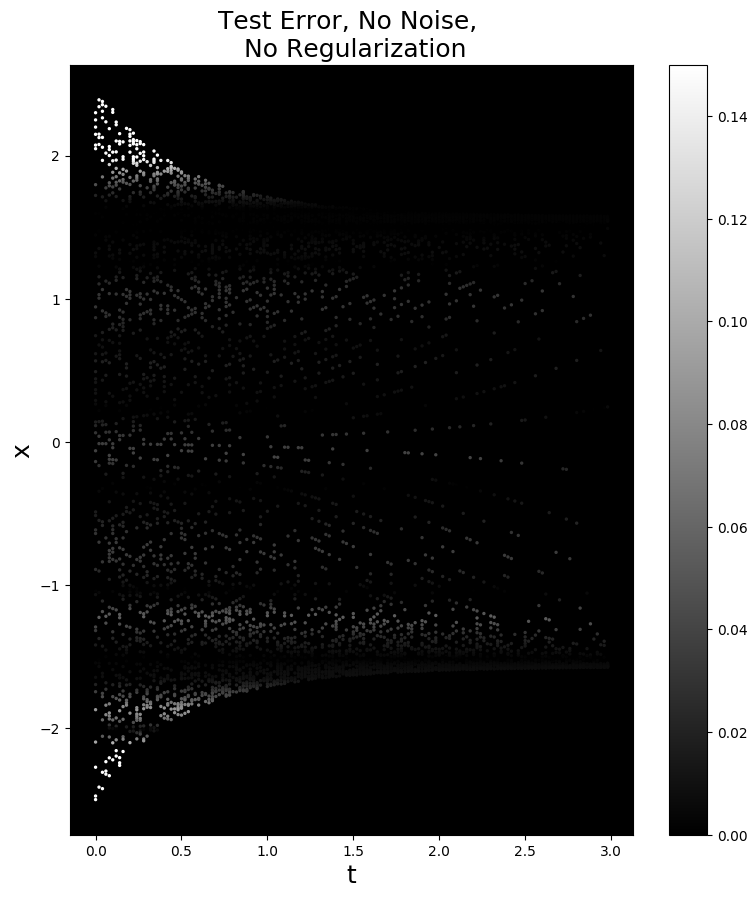}
  \caption{Test error non-regularized network, noiseless data. \vspace{12pt}}
  \label{fig:E0NcosxNR}
\end{subfigure}%
\begin{subfigure}{.5\textwidth}
\captionsetup{width=.8\linewidth}
  \centering
  \includegraphics[width=0.83\linewidth,height=0.85\linewidth,valign=t]{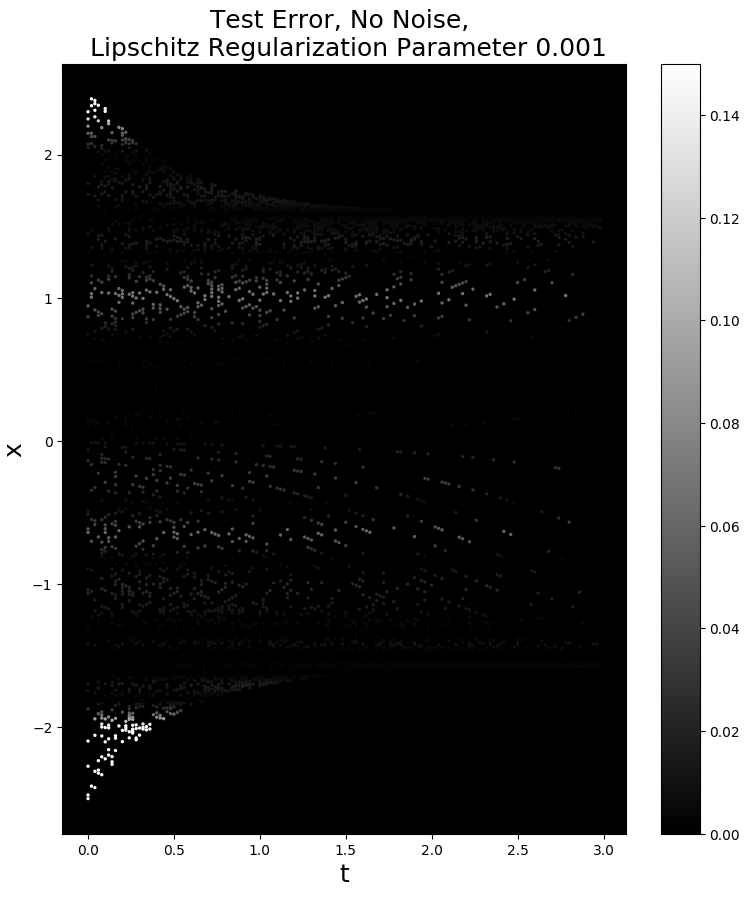}
  \caption{Test error Lipschitz regularized network with parameter 0.001, noiseless data.}
  \label{fig:E0NcosxLR}
\end{subfigure}
\caption{Test error comparison, noiseless data.}
\label{fig:Errxcosx}
\end{figure}

The first numerical example we propose is the recovery on test data of a simple 1-dimensional autonomous ODE: $\dot{x}(t) = x\cos(x)$.\\
The function we are trying to reconstruct on test data is $f(x) = x\cos(x)$.\\
As explained in the data section (Section 2), we generate the data by computing an approximated solution $x(t)$ for the equation using the \texttt{odeint} function in Python. We generate the solution for time steps $t$ linearly spaced in the interval $[0,3]$ with
$\Delta t = 5 \times 10^{-1}$ and for $K = 200$ initial conditions uniformly sampled in the interval $[-2.5,2.5]$. If needed, at this point we add noise to the data.
We then extend the solution by 2 time steps on both sides of the time interval, interpolate the extended solutions using cubic splines and generate target data using difference quotients. For more details on this procedure see Section 2.\\
The hyperparameters of our model are selected in each example by cross validation; in this example, the network $N$ has $L =8$ layers, each layer has 30 neurons; we use minibatches of dimension 50, the learning rate is $10^{-2}$ and it is decreased by a factor of 10 every 7 epochs. We generate results for the regularization parameters: 0 (no regularization), 0.01, 0.005, 0.0025, 0.001.\\
Since our goal is to compare the performance on test data of the networks with an without regularization, we fix a common train MSE across all the regularization parameters choices in order to get a meaningful comparison of the test errors. To do so, we start by training the Lipschitz regularized network with parameter 0.01 for 10 epochs. After the training we obtain what we will be calling a baseline training MSE; for all the other parameter choices, we train the network until the baseline training MSE is reached within an accuracy to the third significant digit.\\
For each choice of the regularization parameter we report the relative baseline train MSE, the relative test MSE and the absolute Generalization Gap. We select the best non-zero regularization parameter as the one that attains the smallest Test error. We compare the regularized and non-regularized network by comparing the corresponding test errors. We also generate plots of the true and recovered right-hand side as well as plots of the absolute errors obtained with the best Lipschitz regularized network and with the non-regularized one. All these results are generated for noiseless data, $1\%$-noisy data and $2\%$-noisy data (for details on how we add noise to the data see Section 2).
\newpage
\noindent
\large{\textbf{Noiseless Data}}\\
\normalsize
We present results in the case of  noiseless data.
\begin{table}[H]
\begin{center}
\begin{tabular}{|
>{\columncolor[HTML]{FFFFFF}}c |
>{\columncolor[HTML]{FFFFFF}}c |
>{\columncolor[HTML]{FFFFFF}}c |
>{\columncolor[HTML]{FFFFFF}}c |}
\hline
{\color[HTML]{000000} \textit{\begin{tabular}[c]{@{}c@{}}Regularization \\ Parameter\end{tabular}}} & {\color[HTML]{000000} \textit{\begin{tabular}[c]{@{}c@{}}Baseline \\ Train MSE\end{tabular}}} & {\color[HTML]{000000} \textit{\begin{tabular}[c]{@{}c@{}}Generalization\\ Gap\end{tabular}}} & {\color[HTML]{000000} \textit{Test MSE}} \\ \hline
{\color[HTML]{000000} \textit{0}}                                                                   & {\color[HTML]{000000} 0.0513\%}                                                               & {\color[HTML]{000000} 2.07e-04}                                                              & {\color[HTML]{000000} 0.0566\%}          \\ \hline
{\color[HTML]{000000} \textit{0.01}}                                                                & {\color[HTML]{000000} 0.0513\%}                                                               & {\color[HTML]{000000} 2.01e-04}                                                              & {\color[HTML]{000000} 0.0564\%}          \\ \hline
{\color[HTML]{000000} \textit{0.005}}                                                               & {\color[HTML]{000000} 0.0513\%}                                                               & {\color[HTML]{000000} 4.30e-05}                                                              & {\color[HTML]{000000} 0.0524\%}          \\ \hline
{\color[HTML]{000000} \textit{0.0025}}                                                              & {\color[HTML]{000000} 0.0513\%}                                                               & {\color[HTML]{000000} 5.16e-05}                                                              & {\color[HTML]{000000} 0.0526\%}          \\ \hline
{\color[HTML]{000000} \textit{\textbf{0.001}}}                                                      & {\color[HTML]{000000} \textbf{0.0513\%}}                                                      & {\color[HTML]{000000} \textbf{3.00e-05}}                                                     & {\color[HTML]{000000} \textbf{0.0521\%}} \\ \hline
\end{tabular}
\captionsetup{width=.8\linewidth}
\caption{Test Error and Generalization Gap comparison for different choices of regularization parameters, no noise in the data.}
\label{tab:cosxT}
\end{center}
\end{table}
\noindent
In Table \ref{tab:cosxT} we report the regularization parameter, the baseline train MSE and test MSE and the generalization gap. We found that, even if all models reach a very good accuracy on test data, the regularized models always generalize better than the non-regularized one and the best accuracy is obtained when adding Lipschitz regularization with parameter $0.001$. This result shows that when no noise is present in the data, there is no need of strong regularization to improve performance.
\medskip\\
We then plot the true and recovered right-hand side and the absolute errors on test data obtained by the best Lipschitz regularized network and by the non-regularized one. In Figure \ref{fig:predxcosx} we represent the test data as a scatter plot in the plane $t,x$ colored according to: the true value of the function $f(t) = x(t)\cos(x(t))$ in figure \ref{fig:Truexcosx}, the predicted value of $f(t)$ when using the non-regularized network in figure \ref{fig:pred0NcosxNR} and the predicted value of $f(t)$ when using the best Lipschitz regularized network in figure \ref{fig:pred0NcosxLR}. We see again that both networks are able to generalize well on test data; in fact the true and recovered functions on test data are nearly indistinguishable. The only small difference can be seen in the upper and lower left corners, where the true function seem to have a larger value (lighter gray level) than in the corresponding predictions.\\
\noindent
In Figure \ref{fig:Errxcosx} we plot the absolute error on test data obtained by the non-regularized and by the best Lipschitz network. The gray level represents the magnitude of the absolute error on a scale from 0 to 0.15, darker gray level means smaller error. From these plots we can clearly see that for both networks the maximum error is attained in the upper and lower left corners. This is expected since by design less training data is available in that part of the domain. We notice also that, as seen in Table \ref{tab:cosxT} the best test MSE is attained by the Lipschitz regularized network. This means that, while for small parts of the domain (for example $x \in [-2.5,-2]$) the error may be larger for the Lipschitz regularized network, the mean error on test data on the whole domain is smaller in the Lipschitz regularized case. Consider now the data-dense subdomain obtained when restricting $x$ to the interval $[-2,2]$. We notice that the Lipschitz regularized network attains a better accuracy overall in this part of the domain. We also notice that the regularized network attains better accuracy than the non-regularized one in the top left corner of the full domain. This shows that the regularized network is able to generalize better than the non-regularized one since it is able to get better accuracy in a part of the domain where less training data was available.

\begin{figure}[H]
\centering
\begin{subfigure}{.5\textwidth}
\captionsetup{width=.8\linewidth}
  \centering
  \includegraphics[width=0.83\linewidth,height=0.85\linewidth,valign=t]{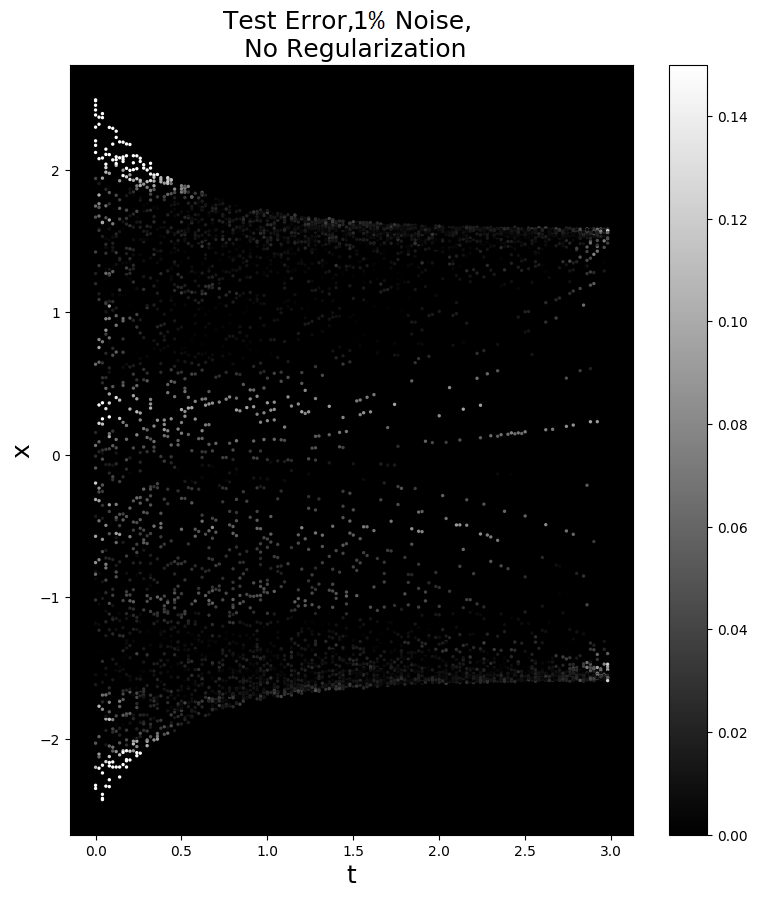}
  \caption{Test error non-regularized network, $1\%$ noisy data. \vspace{12pt}}
  \label{fig:E1NcosxNR}
\end{subfigure}%
\begin{subfigure}{.5\textwidth}
\captionsetup{width=.8\linewidth}
  \centering
  \includegraphics[width=0.82\linewidth,height=0.85\linewidth,valign=t]{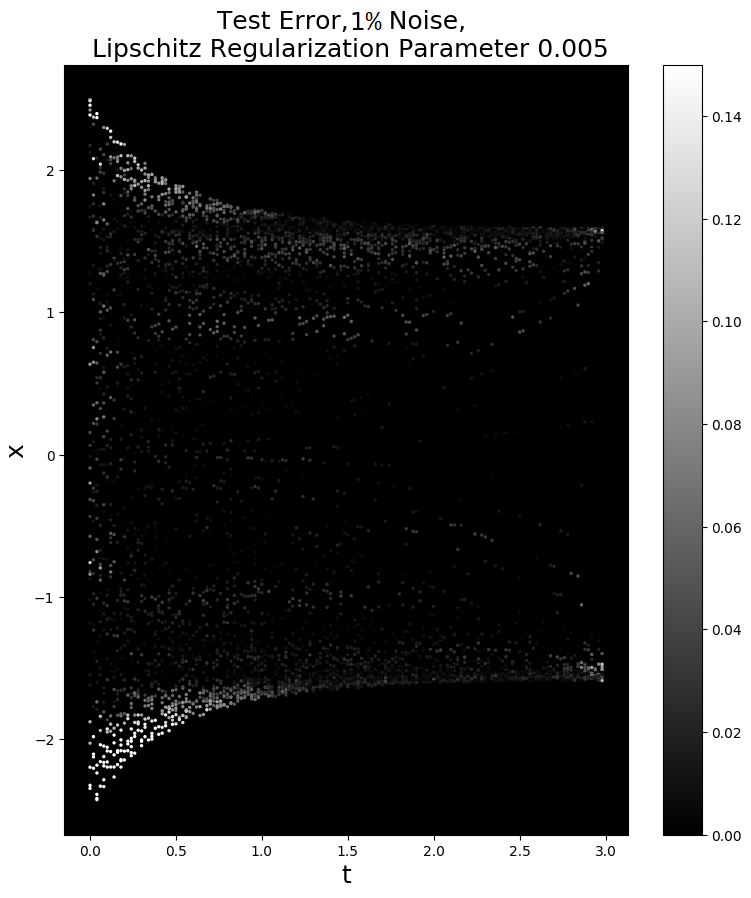}
  \caption{Test error Lipschitz regularized network with parameter 0.005, $1\%$ noisy data.}
  \label{fig:E1NcosxLR}
\end{subfigure}
\caption{Test error comparison, $1\%$ noisy data.}
\label{fig:Err1Nxcosx}
\end{figure}

\begin{figure}[H]
\centering
\begin{subfigure}{.5\textwidth}
\captionsetup{width=.8\linewidth}
  \centering
  \includegraphics[width=0.83\linewidth,height=0.85\linewidth,valign=t]{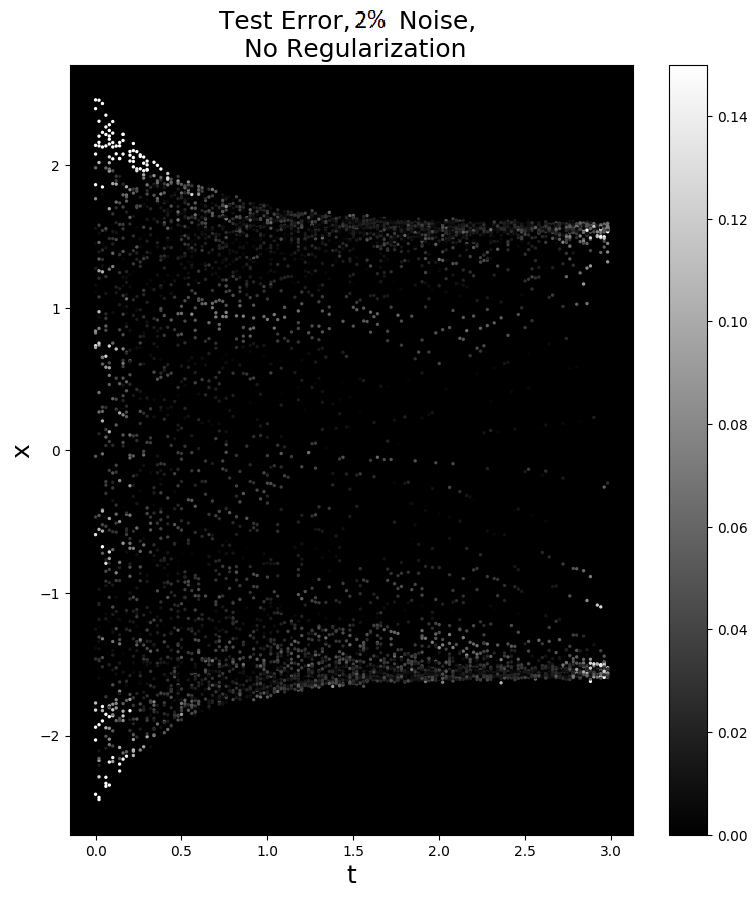}
  \caption{Test error non-regularized network, $2\%$ noisy data. \vspace{12pt}}
  \label{fig:E2NcosxNR}
\end{subfigure}%
\begin{subfigure}{.5\textwidth}
\captionsetup{width=.8\linewidth}
  \centering
  \includegraphics[width=0.83\linewidth, height=0.85\linewidth,valign=t]{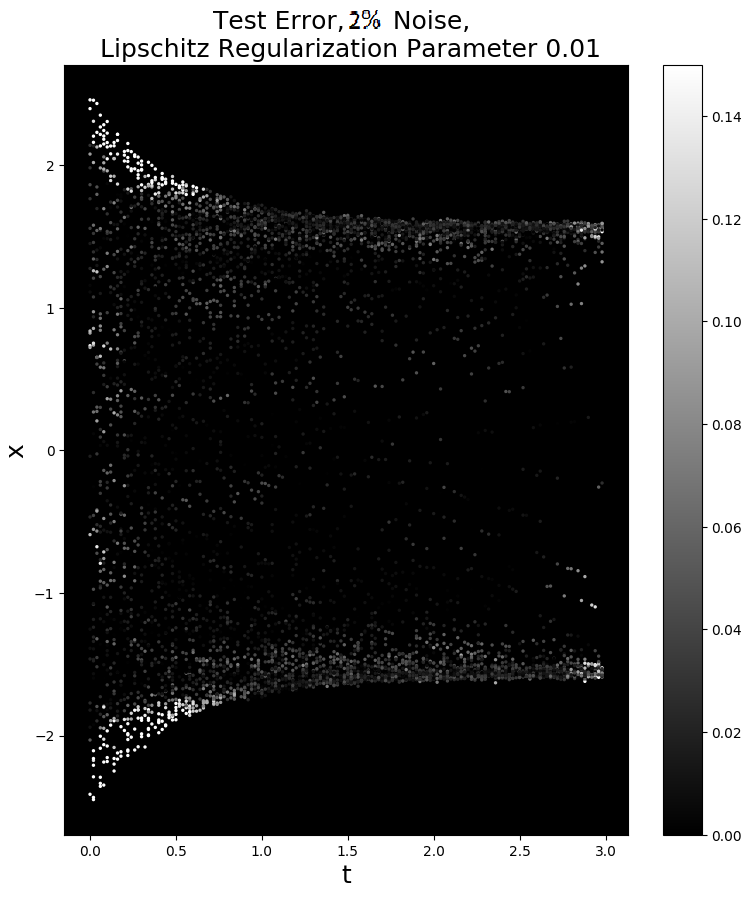}
  \caption{Test error Lipschitz regularized network with parameter 0.01, $2\%$ noisy data.}
  \label{fig:E2NcosxLR}
\end{subfigure}
\caption{Test error comparison, $2\%$ noisy data.}
\label{fig:Err2Nxcosx}
\end{figure}

\bigskip
\noindent
\large{\textbf{Noisy Data}}
\normalsize\\
We present results when adding $1\%$ and $2\%$ of noise to the data.
\begin{table}[H]
\begin{center}
\begin{tabular}{|
>{\columncolor[HTML]{FFFFFF}}c |
>{\columncolor[HTML]{FFFFFF}}c |
>{\columncolor[HTML]{FFFFFF}}c |
>{\columncolor[HTML]{FFFFFF}}c |}
\hline
{\color[HTML]{000000} \textit{\begin{tabular}[c]{@{}c@{}}Regularization \\ Parameter\end{tabular}}} & \multicolumn{1}{c|}{\cellcolor[HTML]{FFFFFF}{\color[HTML]{000000} \textit{\begin{tabular}[c]{@{}c@{}}Baseline \\ Train MSE\end{tabular}}}} & \multicolumn{1}{c|}{\cellcolor[HTML]{FFFFFF}{\color[HTML]{000000} \textit{\begin{tabular}[c]{@{}c@{}}Generalization\\ Gap\end{tabular}}}} & \multicolumn{1}{c|}{\cellcolor[HTML]{FFFFFF}{\color[HTML]{000000} \textit{Test MSE}}} \\ \hline
{\color[HTML]{000000} \textit{0}}                                                                   & {\color[HTML]{000000} 0.0591\%}                                                                                                            & {\color[HTML]{000000} 3.94e-04}                                                                                                           & {\color[HTML]{000000} 0.0693\%}                                                       \\ \hline
{\color[HTML]{000000} \textit{0.01}}                                                                & {\color[HTML]{000000} 0.0591\%}                                                                                                            & {\color[HTML]{000000} 3.05e-04}                                                                                                           & {\color[HTML]{000000} 0.0670\%}                                                       \\ \hline
{\color[HTML]{000000} \textit{\textbf{0.005}}}                                                      & {\color[HTML]{000000} \textbf{0.0591\%}}                                                                                                   & {\color[HTML]{000000} \textbf{1.68e-04}}                                                                                                  & {\color[HTML]{000000} \textbf{0.0635\%}}                                              \\ \hline
{\color[HTML]{000000} \textit{0.0025}}                                                              & {\color[HTML]{000000} 0.0591\%}                                                                                                            & {\color[HTML]{000000} 2.70e-04}                                                                                                           & {\color[HTML]{000000} 0.0661\%}                                                       \\ \hline
{\color[HTML]{000000} \textit{0.001}}                                                               & {\color[HTML]{000000} 0.0591\%}                                                                                                            & {\color[HTML]{000000} 2.54e-04}                                                                                                           & {\color[HTML]{000000} 0.0657\%}                                                       \\ \hline
\end{tabular}
\captionsetup{width=.8\linewidth}
\caption{Test Error and Generalization Gap comparison for different choices of regularization parameters, $1\%$ noise in the data.}
\label{tab:cosxT1N}
\end{center}
\end{table}
\begin{table}[H]
\begin{center}
\begin{tabular}{|
>{\columncolor[HTML]{FFFFFF}}c |
>{\columncolor[HTML]{FFFFFF}}c |
>{\columncolor[HTML]{FFFFFF}}c |
>{\columncolor[HTML]{FFFFFF}}c |}
\hline
{\color[HTML]{000000} \textit{\begin{tabular}[c]{@{}c@{}}Regularization \\ Parameter\end{tabular}}} & {\color[HTML]{000000} \textit{\begin{tabular}[c]{@{}c@{}}Baseline \\ Train MSE\end{tabular}}} & {\color[HTML]{000000} \textit{\begin{tabular}[c]{@{}c@{}}Generalization\\ Gap\end{tabular}}} & {\color[HTML]{000000} \textit{Test MSE}} \\ \hline
{\color[HTML]{000000} \textit{0}}                                                                   & {\color[HTML]{000000} 0.0839\%}                                                               & {\color[HTML]{000000} 3.57e-04}                                                              & {\color[HTML]{000000} 0.0933\%}          \\ \hline
{\color[HTML]{000000} \textit{\textbf{0.01}}}                                                       & {\color[HTML]{000000} \textbf{0.0839\%}}                                                      & {\color[HTML]{000000} \textbf{9.73e-06}}                                                     & {\color[HTML]{000000} \textbf{0.0841\%}} \\ \hline
{\color[HTML]{000000} \textit{0.005}}                                                               & {\color[HTML]{000000} 0.0839\%}                                                               & {\color[HTML]{000000} 3.08e-05}                                                              & {\color[HTML]{000000} 0.0847\%}          \\ \hline
{\color[HTML]{000000} \textit{0.0025}}                                                              & {\color[HTML]{000000} 0.0839\%}                                                               & {\color[HTML]{000000} 5.27e-05}                                                              & {\color[HTML]{000000} 0.0853\%}          \\ \hline
{\color[HTML]{000000} \textit{0.001}}                                                               & {\color[HTML]{000000} 0.0839\%}                                                               & {\color[HTML]{000000} 2.03e-05}                                                              & {\color[HTML]{000000} 0.0844\%}          \\ \hline
\end{tabular}
\captionsetup{width=.8\linewidth}
\caption{Test Error and Generalization Gap comparison for different choices of regularization parameters, $2\%$ noise in the data.}
\label{tab:cosxT2N}
\end{center}
\end{table}
\noindent
In Tables \ref{tab:cosxT1N} and \ref{tab:cosxT2N} we report the training and testing MSE and the generalization gap across all choices of regularization parameters respectively for $1\%$ and $2\%$ of noise in the data. As in the noiseless case, the regularized networks always attain better test accuracy than the non-regularized network, with best test accuracy obtained when the regularization parameter is 0.005 in the $1\%$ noise case and 0.01 in the $2\%$ noise case. These results show that the higher the amount of noise in the data, the larger the regularization parameter which attains best accuracy. This is reasonable since a larger regularization parameter forces the output network to be smoother and thus it prevents overfitting in case of noise in the data. Again the recovery on test data is very good for all choices of the regularization parameters since they all attain a very small relative test error; similar plot as in the noiseless case can be obtained in these cases for the true and predicted function $f$ on test data. We omit these plots.
\medskip\\
We then compare in Figures \ref{fig:Err1Nxcosx} and \ref{fig:Err2Nxcosx} the absolute test error obtained by the non-regularized network and by the best Lipschitz regularized networks. By comparison of Figures \ref{fig:E1NcosxNR} and \ref{fig:E1NcosxLR} and of Figures \ref{fig:E2NcosxNR} and \ref{fig:E2NcosxLR} we can see that in the data-dense subdomain obtained by restricting $x$ to $[-2,2]$ the Lipschitz regularized network attains a better test accuracy than the non-regularized one, showing again that the regularized network is able to generalized better than the non-regularized one. As in the noiseless case, the worst accuracy is attained in the upper and lower left corners of the full domain where less training data was available. We also note that by comparing the noiseless and noisy results, the test error increases as the noise in the data increases, but it is always smaller than $0.1\%$ which shows that all the networks are able to recover this right-hand side function very well even in presence of noise in the data.
\newpage
\subsection{Non-Autonomous 1D ODE} 
To highlight the great applicability of our algorithm, we consider an example in which the right-hand side depends both on space ans time. We propose here the recovery on test data of the ODE:
$$\dot{x}(t) = e^{-x}\log(t) -t^2.$$
The function we are trying to reconstruct is $f(t,x) = e^{-x}\log(t) -t^2$.\\
We generate the solution for time steps $t$ linearly spaced in the interval $[0.1,2]$ with
$\Delta t = 5 \times 10^{-1}$ and for $K = 200$ initial conditions uniformly sampled in the interval $[0.5,5]$. We add noise to the data and generate target data as explained in Section 2.\\
In this example, the network $N$ has $L =8$ layers, each layer has 30 neurons; we use minibatches of dimension 100, the learning rate is $10^{-2}$ and it is decreased by a factor of 10 every 5 epochs. We generate results for the regularization parameters: 0 (no regularization), 0.01, 0.005, 0.0025, 0.001.\\
As in the previous example, we fix a baseline train MSE across all the regularization parameters choices in order to get a meaningful comparison of the test errors.\\
For each choice of the regularization parameter we report the relative baseline train MSE, the relative test MSE and the absolute Generalization Gap. We select the best non-zero regularization parameter as the one that attains the smallest Test error. We compare the regularized and non-regularized network by comparing the corresponding test errors. Since also in this example all the networks attain a very good test accuracy, we omit the plots of the true and predicted function on test data and only show the plots of the absolute errors, since they better show the differences between regularized and non-regularized networks. The results are generated for noiseless data, $1\%$-noisy data and $2\%$-noisy data.
\begin{table}[H]
\begin{center}
\begin{tabular}{|
>{\columncolor[HTML]{FFFFFF}}c |
>{\columncolor[HTML]{FFFFFF}}c |
>{\columncolor[HTML]{FFFFFF}}c |
>{\columncolor[HTML]{FFFFFF}}c |}
\hline
{\color[HTML]{000000} \textit{\begin{tabular}[c]{@{}c@{}}Regularization \\ Parameter\end{tabular}}} & {\color[HTML]{000000} \textit{\begin{tabular}[c]{@{}c@{}}Baseline \\ Train MSE\end{tabular}}} & {\color[HTML]{000000} \textit{\begin{tabular}[c]{@{}c@{}}Generalization\\ Gap\end{tabular}}} & {\color[HTML]{000000} \textit{Test MSE}} \\ \hline
{\color[HTML]{000000} \textit{0}}                                                                   & {\color[HTML]{000000} 0.0521\%}                                                               & {\color[HTML]{000000} 2.85e-04}                                                              & {\color[HTML]{000000} 0.0596\%}          \\ \hline
{\color[HTML]{000000} \textit{0.01}}                                                                & {\color[HTML]{000000} 0.0521\%}                                                               & {\color[HTML]{000000} 2.70e-04}                                                              & {\color[HTML]{000000} 0.0593\%}          \\ \hline
{\color[HTML]{000000} \textit{0.005}}                                                               & {\color[HTML]{000000} 0.0521\%}                                                               & {\color[HTML]{000000} 1.75e-04}                                                              & {\color[HTML]{000000} 0.0564\%}          \\ \hline
{\color[HTML]{000000} \textit{\textbf{0.0025}}}                                                     & {\color[HTML]{000000} \textbf{0.0521\%}}                                                      & {\color[HTML]{000000} \textbf{1.51e-04}}                                                     & {\color[HTML]{000000} \textbf{0.0550\%}} \\ \hline
{\color[HTML]{000000} \textit{0.001}}                                                               & {\color[HTML]{000000} 0.0521\%}                                                               & {\color[HTML]{000000} 1.81e-04}                                                              & {\color[HTML]{000000} 0.0569\%}          \\ \hline
\end{tabular}
\captionsetup{width=.8\linewidth}
\caption{Test Error and Generalization Gap comparison for different choices of regularization parameters, No noise in the data.}
\label{tab:e-xT0N}
\end{center}
\end{table}

\begin{table}[H]
\begin{center}
\begin{tabular}{|
>{\columncolor[HTML]{FFFFFF}}c |
>{\columncolor[HTML]{FFFFFF}}c |
>{\columncolor[HTML]{FFFFFF}}c |
>{\columncolor[HTML]{FFFFFF}}c |}
\hline
{\color[HTML]{000000} \textit{\begin{tabular}[c]{@{}c@{}}Regularization \\ Parameter\end{tabular}}} & {\color[HTML]{000000} \textit{\begin{tabular}[c]{@{}c@{}}Baseline \\ Train MSE\end{tabular}}} & {\color[HTML]{000000} \textit{\begin{tabular}[c]{@{}c@{}}Generalization\\ Gap\end{tabular}}} & {\color[HTML]{000000} \textit{Test MSE}} \\ \hline
{\color[HTML]{000000} \textit{0}}                                                                   & {\color[HTML]{000000} 0.0802\%}                                                               & {\color[HTML]{000000} 3.02e-04}                                                              & {\color[HTML]{000000} 0.0871\%}          \\ \hline
{\color[HTML]{000000} \textit{\textbf{0.01}}}                                                       & {\color[HTML]{000000} \textbf{0.0802\%}}                                                      & {\color[HTML]{000000} \textbf{1.05e-04}}                                                     & {\color[HTML]{000000} \textbf{0.0826\%}} \\ \hline
{\color[HTML]{000000} \textit{0.005}}                                                               & {\color[HTML]{000000} 0.0802\%}                                                               & {\color[HTML]{000000} 2.64e-04}                                                              & {\color[HTML]{000000} 0.0863\%}          \\ \hline
{\color[HTML]{000000} \textit{0.0025}}                                                              & {\color[HTML]{000000} 0.0802\%}                                                               & {\color[HTML]{000000} 1.86e-04}                                                              & {\color[HTML]{000000} 0.0845\%}          \\ \hline
{\color[HTML]{000000} \textit{0.001}}                                                               & {\color[HTML]{000000} 0.0802\%}                                                               & {\color[HTML]{000000} 2.05e-04}                                                              & {\color[HTML]{000000} 0.0849\%}          \\ \hline
\end{tabular}
\captionsetup{width=.8\linewidth}
\caption{Test Error and Generalization Gap comparison for different choices of regularization parameters, $1\%$ noise in the data.}
\label{tab:e-xT1N}
\end{center}
\end{table}

\begin{figure}[H]
\centering
\begin{subfigure}{.5\textwidth}
\captionsetup{width=.8\linewidth}
  \centering
  \includegraphics[width=0.85\linewidth,height=0.85\linewidth,valign=t]{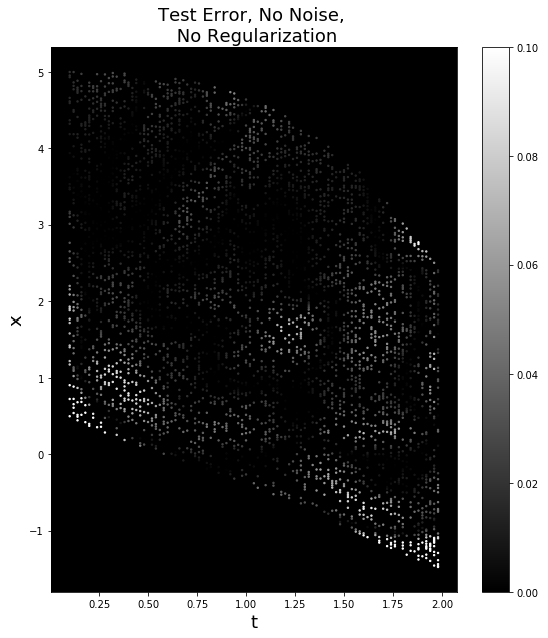}
  \caption{Test error non-regularized network, No noise in the data. \vspace{12pt}}
  \label{fig:E0e-xNR}
\end{subfigure}%
\begin{subfigure}{.5\textwidth}
\captionsetup{width=.8\linewidth}
  \centering
  \includegraphics[width=0.83\linewidth,height=0.85\linewidth,valign=t]{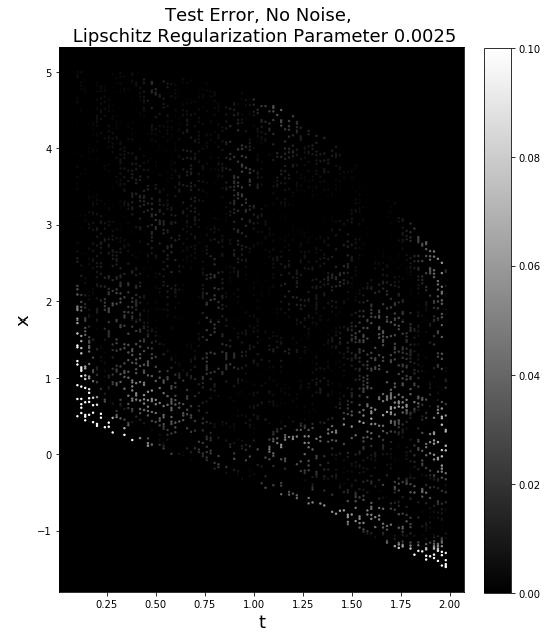}
  \caption{Test error Lipschitz regularized network with parameter 0.0025, no noise in the data.}
  \label{fig:E0e-xLR}
\end{subfigure}
\caption{Test error comparison,$1\%$ noisy data.}
\label{fig:Err0Ne-x}
\end{figure}

\begin{figure}[H]
\centering
\begin{subfigure}{.5\textwidth}
\captionsetup{width=.8\linewidth}
  \centering
  \includegraphics[width=0.85\linewidth,height=0.85\linewidth,valign=t]{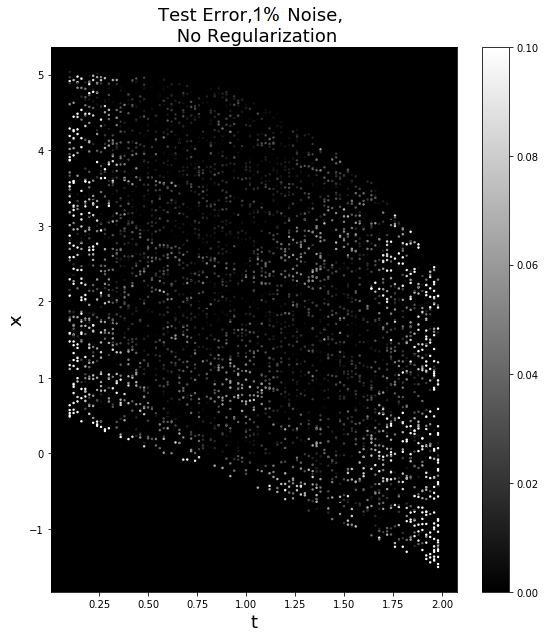}
  \caption{Test error non-regularized network, $1\%$ noisy data. \vspace{12pt}}
  \label{fig:E1Ne-xNR}
\end{subfigure}%
\begin{subfigure}{.5\textwidth}
\captionsetup{width=.8\linewidth}
  \centering
  \includegraphics[width=0.83\linewidth, height=0.85\linewidth,valign=t]{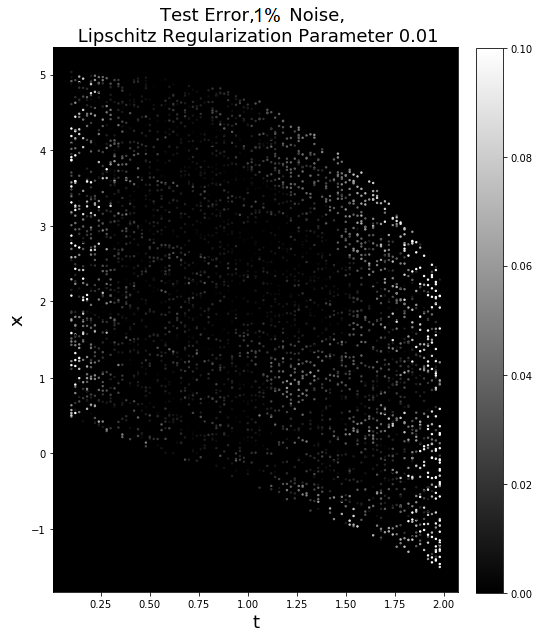}
  \caption{Test error Lipschitz regularized network with parameter 0.01, $1\%$ noisy data.}
  \label{fig:E1Ne-xLR}
\end{subfigure}
\caption{Test error comparison, $1\%$ noisy data.}
\label{fig:Err1Ne-x}
\end{figure}

\begin{figure}[H]
\centering
\begin{subfigure}{.5\textwidth}
\captionsetup{width=.8\linewidth}
  \centering
  \includegraphics[width=0.85\linewidth,height=0.85\linewidth, valign=t]{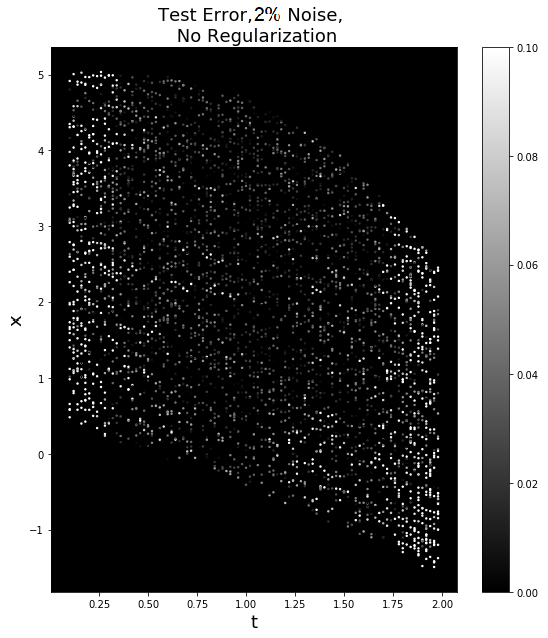}
  \caption{Test error non-regularized network, $2\%$ noisy data.\vspace{12pt}}
  \label{fig:E2Ne-xNR}
\end{subfigure}%
\begin{subfigure}{.5\textwidth}
\captionsetup{width=.8\linewidth}
  \centering
  \includegraphics[width=0.83\linewidth, height=0.85\linewidth, valign=t]{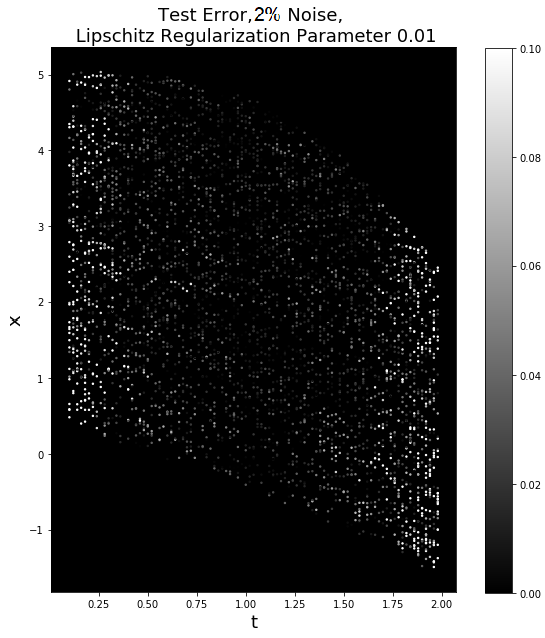}
  \caption{Test error Lipschitz regularized network with parameter 0.01, $2\%$ noisy data.}
  \label{fig:E2Ne-xLR}
\end{subfigure}
\caption{Test error comparison, $2\%$ noisy data.}
\label{fig:Err2Ne-x}
\end{figure}

\begin{table}[H]
\begin{center}
\begin{tabular}{|
>{\columncolor[HTML]{FFFFFF}}c |
>{\columncolor[HTML]{FFFFFF}}c |
>{\columncolor[HTML]{FFFFFF}}c |
>{\columncolor[HTML]{FFFFFF}}c |}
\hline
{\color[HTML]{000000} \textit{\begin{tabular}[c]{@{}c@{}}Regularization \\ Parameter\end{tabular}}} & {\color[HTML]{000000} \textit{\begin{tabular}[c]{@{}c@{}}Baseline \\ Train MSE\end{tabular}}} & {\color[HTML]{000000} \textit{\begin{tabular}[c]{@{}c@{}}Generalization\\ Gap\end{tabular}}} & {\color[HTML]{000000} \textit{Test MSE}} \\ \hline
{\color[HTML]{000000} \textit{0}}                                                                   & {\color[HTML]{000000} 0.183\%}                                                               & {\color[HTML]{000000} 5.78e-04}                                                              & {\color[HTML]{000000} 0.195\%}          \\ \hline
{\color[HTML]{000000} \textit{\textbf{0.01}}}                                                       & {\color[HTML]{000000} \textbf{0.1832\%}}                                                      & {\color[HTML]{000000} \textbf{4.89e-04}}                                                     & {\color[HTML]{000000} \textbf{0.193\%}} \\ \hline
{\color[HTML]{000000} \textit{0.005}}                                                               & {\color[HTML]{000000} 0.1832\%}                                                               & {\color[HTML]{000000} 5.35e-04}                                                              & {\color[HTML]{000000} 0.194\%}          \\ \hline
{\color[HTML]{000000} \textit{0.0025}}                                                              & {\color[HTML]{000000} 0.1832\%}                                                               & {\color[HTML]{000000} 6.08e-04}                                                              & {\color[HTML]{000000} 0.196\%}          \\ \hline
{\color[HTML]{000000} \textit{0.001}}                                                               & {\color[HTML]{000000} 0.1832\%}                                                               & {\color[HTML]{000000} 5.22e-04}                                                              & {\color[HTML]{000000} 0.194\%}          \\ \hline
\end{tabular}
\captionsetup{width=.8\linewidth}
\caption{Test Error and Generalization Gap comparison for different choices of regularization parameters, $2\%$ noise in the data.}
\label{tab:e-xT2N}
\end{center}
\end{table}
\noindent
In Tables \ref{tab:e-xT0N}, \ref{tab:e-xT1N} and \ref{tab:e-xT2N} we report the regularization parameter, the baseline train MSE and test MSE and the generalization gap respectively in the case of noiseless data $1\%$ and $2\%$-noisy data. We found that, even if all models reach a very good accuracy on test data, the regularized models in general generalize better than the non-regularized one and the best accuracy is obtained when adding Lipschitz regularization with parameter $0.0025$ in the noiseless case, $0.01$ in both the noisy cases presented. As in the previous example we see that when no noise is present in the data, a small regularization parameter attains the best test error, while when the data is noisy, a larger regularization parameter helps generalization since it prevents overfitting by forcing the network to be smoother. We note though that even in the case of noiseless data, adding a Lipschitz regularization term improves generalization. Finally we note that in all cases the test error is  smaller than $0.2\%$ which shows that all the networks are able to recover this right-hand side function very well even in presence of noise in the data.
\medskip\\
In Figures \ref{fig:Err0Ne-x}, \ref{fig:Err1Ne-x} and  \ref{fig:Err2Ne-x} we compare the absolute test error obtained by the non-regularized network and by the best Lipschitz regularized networks. The gray level represents the magnitude of the absolute error on a scale from 0 to 0.10, darker gray level means smaller error. Comparing these three figures we can see how the test error increases as we increase the amount of noise in the data.  By comparison of Figures \ref{fig:E0e-xNR} and \ref{fig:E0e-xLR}, of Figures \ref{fig:E1Ne-xNR} and \ref{fig:E1Ne-xLR} and of Figures \ref{fig:E2Ne-xNR} and \ref{fig:E2Ne-xLR}, we can see that largest errors for all networks are attained at the boundaries of the temporal domain, this is reasonable since for a central point the network can use neighboring points to better estimate the function value, while for boundary points this is not possible since there is no data for $t<0$ and $t>2$. The best accuracy is  is attained in the central part of the domain where the Lipschitz regularized networks perform best, showing again that regularized networks are able to generalized better than the non-regularized one. 

\newpage
\subsection{Autonomous ODE System}
We present here a higher dimensional dimensional example in which we recover a system of equations. We propose the recovery on test data the following Lotka-Volterra system of differential equations:
\begin{align*}
    \begin{cases}
     \dot{x}_1 = 1.5x_1 - x_1x_2;\\
    \dot{x}_2 = -3x_2 +x_1x_2.
    \end{cases}
\end{align*}
The functions we are trying to reconstruct are: $$f_1(x_1,x_2) = 1.5x_1 - x_1x_2, \quad f_2(x_1,x_2) = -3x_2 +x_1x_2.$$
The recovery is done separately for the two components $f_1$ and $f_2$ each of which is approximated by a neural network. We note that, given the similar form of $f_1$ and $f_2$, it is possible to use only one network to approximate both components, however in this paper we assume no knowledge on the form of the right-hand side, so we recover the two components using two different networks.
We generate the solution for time steps $t$ linearly spaced in the interval $[0,4]$ with
$\Delta t = 5 \times 10^{-1}$ and for $K = 400$ initial conditions uniformly sampled in the interval $[1,5]$. We add noise to the data and generate reliable target data as explained in Section 2. We note that in this case we are working in a 2-dimensional space; because of the curse of dimensionality, in order to obtain good accuracy we need more data than in the 1-dimensionl examples.\\
In this example, the networks $N_1$ and $N_2$ that approximate the two components of the right-hand side  have $L =10$ layers and each layer has 50 neurons; we use minibatches of dimension 200, the learning rate is $10^{-2}$ and it is decreased by a factor of 10 every 3 epochs. We generate results for the regularization parameters: 0 (no regularization), 0.01, 0.005, 0.0025, 0.001.
\begin{table}[H]
\begin{center}
\resizebox{\columnwidth}{!}{%
\begin{tabular}{|
>{\columncolor[HTML]{FFFFFF}}c |
>{\columncolor[HTML]{FFFFFF}}c |
>{\columncolor[HTML]{FFFFFF}}c |
>{\columncolor[HTML]{FFFFFF}}c |
>{\columncolor[HTML]{FFFFFF}}c |
>{\columncolor[HTML]{FFFFFF}}c |
>{\columncolor[HTML]{FFFFFF}}c |
>{\columncolor[HTML]{FFFFFF}}c |}
\hline
\multicolumn{4}{|c|}{\cellcolor[HTML]{FFFFFF}{\color[HTML]{000000} \textit{First Component}}}                                                                                                                                                                                                                                               & \multicolumn{4}{c|}{\cellcolor[HTML]{FFFFFF}{\color[HTML]{000000} \textit{Second Component}}}                                                                                                                                                                                                                                               \\ \hline
{\color[HTML]{000000} \textit{\begin{tabular}[c]{@{}c@{}}Regularization\\ Parameter\end{tabular}}} & {\color[HTML]{000000} \textit{\begin{tabular}[c]{@{}c@{}}Baseline\\ Train MSE\end{tabular}}} & {\color[HTML]{000000} \textit{\begin{tabular}[c]{@{}c@{}}Generalization\\ Gap\end{tabular}}} & {\color[HTML]{000000} \textit{Test MSE}} & {\color[HTML]{000000} \textit{\begin{tabular}[c]{@{}c@{}}Regularization\\ Parameter\end{tabular}}} & {\color[HTML]{000000} \textit{\begin{tabular}[c]{@{}c@{}}Baseline\\ Train MSE\end{tabular}}} & {\color[HTML]{000000} \textit{\begin{tabular}[c]{@{}c@{}}Generalization\\ Gap\end{tabular}}} & {\color[HTML]{000000} \textit{Test MSE}} \\ \hline
{\color[HTML]{000000} \textit{0}}                                                                  & {\color[HTML]{000000} 0.0120\%}                                                              & {\color[HTML]{000000} 2.10e-04}                                                              & {\color[HTML]{000000} 0.0128\%}          & {\color[HTML]{000000} \textit{0}}                                                                  & {\color[HTML]{000000} 0.0160\%}                                                              & {\color[HTML]{000000} 3.07e-04}                                                              & {\color[HTML]{000000} 0.0174\%}          \\ \hline
{\color[HTML]{000000} \textit{0.01}}                                                               & {\color[HTML]{000000} 0.0120\%}                                                              & {\color[HTML]{000000} 8.34e-05}                                                              & {\color[HTML]{000000} 0.0123\%}          & {\color[HTML]{000000} \textit{0.01}}                                                               & {\color[HTML]{000000} 0.0160\%}                                                              & {\color[HTML]{000000} 8.67e-05}                                                              & {\color[HTML]{000000} 0.0164\%}          \\ \hline
{\color[HTML]{000000} \textit{0.005}}                                                              & {\color[HTML]{000000} 0.0120\%}                                                              & {\color[HTML]{000000} 1.35e-04}                                                              & {\color[HTML]{000000} 0.0125\%}          & {\color[HTML]{000000} \textit{0.005}}                                                              & {\color[HTML]{000000} 0.0160\%}                                                              & {\color[HTML]{000000} 4.76e-05}                                                              & {\color[HTML]{000000} 0.0162\%}          \\ \hline
{\color[HTML]{000000} \textit{0.0025}}                                                             & {\color[HTML]{000000} 0.0120\%}                                                              & {\color[HTML]{000000} 1.56e-04}                                                              & {\color[HTML]{000000} 0.0126\%}          & {\color[HTML]{000000} \textit{0.0025}}                                                             & {\color[HTML]{000000} 0.0160\%}                                                              & {\color[HTML]{000000} 1.13e-04}                                                              & {\color[HTML]{000000} 0.0165\%}          \\ \hline
{\color[HTML]{000000} \textit{\textbf{0.001}}}                                                     & {\color[HTML]{000000} \textbf{0.0120\%}}                                                     & {\color[HTML]{000000} \textbf{3.11e-05}}                                                     & {\color[HTML]{000000} \textbf{0.0121\%}} & {\color[HTML]{000000} \textit{\textbf{0.001}}}                                                     & {\color[HTML]{000000} \textbf{0.0160\%}}                                                     & {\color[HTML]{000000} \textbf{2.95e-05}}                                                     & {\color[HTML]{000000} \textbf{0.0161\%}} \\ \hline
\end{tabular}%
}
\captionsetup{width=.8\linewidth}
\caption{Test Error and Generalization Gap comparison for different choices of regularization parameters,  No noise in the data.}
\label{tab:LVT0N}
\end{center}
\end{table}

\begin{table}[H]
\begin{center}
\resizebox{\columnwidth}{!}{%
\begin{tabular}{|
>{\columncolor[HTML]{FFFFFF}}c |
>{\columncolor[HTML]{FFFFFF}}c |
>{\columncolor[HTML]{FFFFFF}}c |
>{\columncolor[HTML]{FFFFFF}}c |
>{\columncolor[HTML]{FFFFFF}}c |
>{\columncolor[HTML]{FFFFFF}}c |
>{\columncolor[HTML]{FFFFFF}}c |
>{\columncolor[HTML]{FFFFFF}}c |}
\hline
\multicolumn{4}{|c|}{\cellcolor[HTML]{FFFFFF}{\color[HTML]{000000} \textit{First Component}}}                                                                                                                                                                                                                                               & \multicolumn{4}{c|}{\cellcolor[HTML]{FFFFFF}{\color[HTML]{000000} \textit{Second Component}}}                                                                                                                                                                                                                                               \\ \hline
{\color[HTML]{000000} \textit{\begin{tabular}[c]{@{}c@{}}Regularization\\ Parameter\end{tabular}}} & {\color[HTML]{000000} \textit{\begin{tabular}[c]{@{}c@{}}Baseline\\ Train MSE\end{tabular}}} & {\color[HTML]{000000} \textit{\begin{tabular}[c]{@{}c@{}}Generalization\\ Gap\end{tabular}}} & {\color[HTML]{000000} \textit{Test MSE}} & {\color[HTML]{000000} \textit{\begin{tabular}[c]{@{}c@{}}Regularization\\ Parameter\end{tabular}}} & {\color[HTML]{000000} \textit{\begin{tabular}[c]{@{}c@{}}Baseline\\ Train MSE\end{tabular}}} & {\color[HTML]{000000} \textit{\begin{tabular}[c]{@{}c@{}}Generalization\\ Gap\end{tabular}}} & {\color[HTML]{000000} \textit{Test MSE}} \\ \hline
{\color[HTML]{000000} \textit{0}}                                                                  & {\color[HTML]{000000} 0.238\%}                                                               & {\color[HTML]{000000} 1.34e-03}                                                              & {\color[HTML]{000000} 0.244\%}           & {\color[HTML]{000000} \textit{0}}                                                                  & {\color[HTML]{000000} 0.266\%}                                                               & {\color[HTML]{000000} 3.23e-03}                                                              & {\color[HTML]{000000} 0.280\%}           \\ \hline
{\color[HTML]{000000} \textit{0.01}}                                                               & {\color[HTML]{000000} 0.238\%}                                                               & {\color[HTML]{000000} 3.86e-04}                                                              & {\color[HTML]{000000} 0.239\%}           & {\color[HTML]{000000} \textit{0.01}}                                                               & {\color[HTML]{000000} 0.266\%}                                                               & {\color[HTML]{000000} 1.50e-03}                                                              & {\color[HTML]{000000} 0.272\%}           \\ \hline
{\color[HTML]{000000} \textit{\textbf{0.005}}}                                                     & {\color[HTML]{000000} \textbf{0.238\%}}                                                      & {\color[HTML]{000000} \textbf{3.36e-04}}                                                     & {\color[HTML]{000000} \textbf{0.239\%}}  & {\color[HTML]{000000} \textit{\textbf{0.005}}}                                                     & {\color[HTML]{000000} \textbf{0.266\%}}                                                      & {\color[HTML]{000000} \textbf{4.83e-04}}                                                     & {\color[HTML]{000000} \textbf{0.268\%}}  \\ \hline
{\color[HTML]{000000} \textit{0.0025}}                                                             & {\color[HTML]{000000} 0.238\%}                                                               & {\color[HTML]{000000} 1.06e-03}                                                              & {\color[HTML]{000000} 0.243\%}           & {\color[HTML]{000000} \textit{0.0025}}                                                             & {\color[HTML]{000000} 0.266\%}                                                               & {\color[HTML]{000000} 1.70e-03}                                                              & {\color[HTML]{000000} 0.273\%}           \\ \hline
{\color[HTML]{000000} \textit{0.001}}                                                              & {\color[HTML]{000000} 0.238\%}                                                               & {\color[HTML]{000000} 1.16e-03}                                                              & {\color[HTML]{000000} 0.243\%}           & {\color[HTML]{000000} \textit{0.001}}                                                              & {\color[HTML]{000000} 0.266\%}                                                               & {\color[HTML]{000000} 1.52e-03}                                                              & {\color[HTML]{000000} 0.273\%}           \\ \hline
\end{tabular}%
}
\captionsetup{width=.8\linewidth}
\caption{Test Error and Generalization Gap comparison for different choices of regularization parameters,  $1\%$ noise in the data.}
\label{tab:LVT1N}
\end{center}
\end{table}

\begin{table}[H]
\begin{center}
\resizebox{\columnwidth}{!}{%
\begin{tabular}{|
>{\columncolor[HTML]{FFFFFF}}c |
>{\columncolor[HTML]{FFFFFF}}c |
>{\columncolor[HTML]{FFFFFF}}c |
>{\columncolor[HTML]{FFFFFF}}c |
>{\columncolor[HTML]{FFFFFF}}c |
>{\columncolor[HTML]{FFFFFF}}c |
>{\columncolor[HTML]{FFFFFF}}c |
>{\columncolor[HTML]{FFFFFF}}c |}
\hline
\multicolumn{4}{|c|}{\cellcolor[HTML]{FFFFFF}{\color[HTML]{000000} \textit{First Component}}}                                                                                                                                                                                                                                               & \multicolumn{4}{c|}{\cellcolor[HTML]{FFFFFF}{\color[HTML]{000000} \textit{Second Component}}}                                                                                                                                                                                                                                               \\ \hline
{\color[HTML]{000000} \textit{\begin{tabular}[c]{@{}c@{}}Regularization\\ Parameter\end{tabular}}} & {\color[HTML]{000000} \textit{\begin{tabular}[c]{@{}c@{}}Baseline\\ Train MSE\end{tabular}}} & {\color[HTML]{000000} \textit{\begin{tabular}[c]{@{}c@{}}Generalization\\ Gap\end{tabular}}} & {\color[HTML]{000000} \textit{Test MSE}} & {\color[HTML]{000000} \textit{\begin{tabular}[c]{@{}c@{}}Regularization\\ Parameter\end{tabular}}} & {\color[HTML]{000000} \textit{\begin{tabular}[c]{@{}c@{}}Baseline\\ Train MSE\end{tabular}}} & {\color[HTML]{000000} \textit{\begin{tabular}[c]{@{}c@{}}Generalization\\ Gap\end{tabular}}} & {\color[HTML]{000000} \textit{Test MSE}} \\ \hline
{\color[HTML]{000000} \textit{0}}                                                                  & {\color[HTML]{000000} 0.544\%}                                                               & {\color[HTML]{000000} 1.21e-02}                                                              & {\color[HTML]{000000} 0.587\%}           & {\color[HTML]{000000} \textit{0}}                                                                  & {\color[HTML]{000000} 0.509\%}                                                               & {\color[HTML]{000000} 5.51e-03}                                                              & {\color[HTML]{000000} 0.533\%}           \\ \hline
{\color[HTML]{000000} \textit{\textbf{0.01}}}                                                      & {\color[HTML]{000000} \textbf{0.544\%}}                                                      & {\color[HTML]{000000} \textbf{6.26e-03}}                                                     & {\color[HTML]{000000} \textbf{0.567\%}}  & {\color[HTML]{000000} \textit{\textbf{0.01}}}                                                      & {\color[HTML]{000000} \textbf{0.509\%}}                                                      & {\color[HTML]{000000} \textbf{9.45e-04}}                                                     & {\color[HTML]{000000} \textbf{0.513\%}}  \\ \hline
{\color[HTML]{000000} \textit{0.005}}                                                              & {\color[HTML]{000000} 0.544\%}                                                               & {\color[HTML]{000000} 7.48e-03}                                                              & {\color[HTML]{000000} 0.571\%}           & {\color[HTML]{000000} \textit{0.005}}                                                              & {\color[HTML]{000000} 0.509\%}                                                               & {\color[HTML]{000000} 1.08e-03}                                                              & {\color[HTML]{000000} 0.513\%}           \\ \hline
{\color[HTML]{000000} \textit{0.0025}}                                                             & {\color[HTML]{000000} 0.544\%}                                                               & {\color[HTML]{000000} 6.50e-03}                                                              & {\color[HTML]{000000} 0.568\%}           & {\color[HTML]{000000} \textit{0.0025}}                                                             & {\color[HTML]{000000} 0.509\%}                                                               & {\color[HTML]{000000} 1.36e-03}                                                              & {\color[HTML]{000000} 0.515\%}           \\ \hline
{\color[HTML]{000000} \textit{0.001}}                                                              & {\color[HTML]{000000} 0.544\%}                                                               & {\color[HTML]{000000} 6.86e-03}                                                              & {\color[HTML]{000000} 0.569\%}           & {\color[HTML]{000000} \textit{0.001}}                                                              & {\color[HTML]{000000} 0.509\%}                                                               & {\color[HTML]{000000} 1.14e-03}                                                              & {\color[HTML]{000000} 0.514\%}           \\ \hline
\end{tabular}%
}
\captionsetup{width=.8\linewidth}
\caption{Test Error and Generalization Gap comparison for different choices of regularization parameters,  $2\%$ noise in the data.}
\label{tab:LVT2N}
\end{center}
\end{table}
\noindent
In Tables \ref{tab:LVT0N}, \ref{tab:LVT1N} and \ref{tab:LVT2N} we report the regularization parameter, the baseline train MSE and test MSE and the generalization gap for both components, respectively in the case of noiseless data $1\%$ and $2\%$-noisy data. 
The recovery is very good for all networks, with regularized network performing better in all cases. For clarity, when the same relative Test MSE is attained by two models, the best model is the one which has the smaller generalization gap. We note again that when more noise is present in the data, a larger regularization parameter is preferable since it gives better test accuracy. Specifically, when no noise is added to the data, the best regularization parameter is the 0.001, while when we add $1\%$ and $2\%$ of noise in the data the best parameters are respectively 0.005 and 0.01. As we noted above, since the functional form of $f_1$ and $f_2$ is very similar, we obtain very similar accuracy for both components. In all cases, the test error is  smaller than $0.6\%$.
\medskip\\
As in the previous two examples, we generated plots of the absolute errors on test data obtained by the non-regularized and best Lipschitz regularized networks. Similar results as in previous cases are observed: an overall better accuracy obtained by the Lipschitz regularized network for all amounts of noise in the data and largest errors attained at the boundaries of the domain and where the training data was sparse. For brevity we omit these plots here, but they can be found in the additional material section of the paper.
\subsection{Second Order Non-Autonomous ODE}
FInally, in the last example we consider a non autonomous differential equation of second order. This can be reduced to a system of two ODEs of first order having both space and time dependence in the right-hand side as well as an oscillatory term which may make the recovery harder.\\ We propose the recovery on test data of the following pendulum equation:
$$\ddot{z} + 2\dot{z} +2z = \cos(2t)$$
This can be rewritten in the usual way as a system of two first order ODEs by setting $x_1 := z, x_2:=\dot{z}$. We obtain:
\begin{align*}
    \begin{cases}
    \dot{x}_1 = x_2\\
    \dot{x}_2 = -2x_1 -2x_2 + \cos(2t)
    \end{cases}
\end{align*}
The functions we are trying to reconstruct on test data are: $$f_1(t, x_1,x_2) =x_2, \quad f_2(t, x_1,x_2) = -2x_1 -2x_2 + \cos(2t)$$
In this case the recovery is done separately for the two components $f_1$ and $f_2$ each of which is approximated by a neural network.
We generate the solution for time steps $t$ linearly spaced in the interval $[0,2]$ with
$\Delta t = 5 \times 10^{-1}$ and for $K = 1000$ initial conditions uniformly sampled in the interval $[0,2]$. We add noise to the data and generate reliable target data as explained in Section 2. We note that in this case we are working in a 3-dimensional space; because of the curse of dimensionality, in order to obtain good accuracy we need more data than in the 1-dimensional examples.\\
In this example, the networks $N_1$ and $N_2$ that approximate the two components of the right-hand side  have $L =10$ layers and each layer has 60 neurons; we use minibatches of dimension 100, the learning rate is $10^{-2}$ and it is decreased by a factor of 10 every 3 epochs. We generate results for the regularization parameters: 0 (no regularization), 0.01, 0.005, 0.0025, 0.001. We fix a baseline train MSE across all the regularization parameters, we report the relative test MSE and the absolute Generalization Gap, and select the best regularization parameter for both components.
\begin{table}[H]
\begin{center}
\resizebox{\columnwidth}{!}{%
\begin{tabular}{|
>{\columncolor[HTML]{FFFFFF}}c |
>{\columncolor[HTML]{FFFFFF}}c |
>{\columncolor[HTML]{FFFFFF}}c |
>{\columncolor[HTML]{FFFFFF}}c |
>{\columncolor[HTML]{FFFFFF}}c |
>{\columncolor[HTML]{FFFFFF}}c |
>{\columncolor[HTML]{FFFFFF}}c |
>{\columncolor[HTML]{FFFFFF}}c |}
\hline
\multicolumn{4}{|c|}{\cellcolor[HTML]{FFFFFF}{\color[HTML]{000000} \textit{First Component}}}                                                                                                                                                                                                                                                & \multicolumn{4}{c|}{\cellcolor[HTML]{FFFFFF}{\color[HTML]{000000} \textit{Second Component}}}                                                                                                                                                                                                                                               \\ \hline
{\color[HTML]{000000} \textit{\begin{tabular}[c]{@{}c@{}}Regularization\\ Parameter\end{tabular}}} & {\color[HTML]{000000} \textit{\begin{tabular}[c]{@{}c@{}}Baseline\\ Train MSE\end{tabular}}} & {\color[HTML]{000000} \textit{\begin{tabular}[c]{@{}c@{}}Generalization\\ Gap\end{tabular}}} & {\color[HTML]{000000} \textit{Test MSE}}  & {\color[HTML]{000000} \textit{\begin{tabular}[c]{@{}c@{}}Regularization\\ Parameter\end{tabular}}} & {\color[HTML]{000000} \textit{\begin{tabular}[c]{@{}c@{}}Baseline\\ Train MSE\end{tabular}}} & {\color[HTML]{000000} \textit{\begin{tabular}[c]{@{}c@{}}Generalization\\ Gap\end{tabular}}} & {\color[HTML]{000000} \textit{Test MSE}} \\ \hline
{\color[HTML]{000000} \textit{0}}                                                                  & {\color[HTML]{000000} 0.00829\%}                                                             & {\color[HTML]{000000} 8.71e-06}                                                              & {\color[HTML]{000000} 0.00854\%}          & {\color[HTML]{000000} \textit{0}}                                                                  & {\color[HTML]{000000} 0.0103\%}                                                              & {\color[HTML]{000000} 7.18e-06}                                                              & {\color[HTML]{000000} 0.0104\%}          \\ \hline
{\color[HTML]{000000} \textit{0.01}}                                                               & {\color[HTML]{000000} 0.00829\%}                                                             & {\color[HTML]{000000} 4.23e-06}                                                              & {\color[HTML]{000000} 0.00841\%}          & {\color[HTML]{000000} \textit{0.01}}                                                               & {\color[HTML]{000000} 0.0103\%}                                                              & {\color[HTML]{000000} 4.23e-06}                                                              & {\color[HTML]{000000} 0.0103\%}          \\ \hline
{\color[HTML]{000000} \textit{0.005}}                                                              & {\color[HTML]{000000} 0.00829\%}                                                             & {\color[HTML]{000000} 5.80e-06}                                                              & {\color[HTML]{000000} 0.00845\%}          & {\color[HTML]{000000} \textit{0.005}}                                                              & {\color[HTML]{000000} 0.0103\%}                                                              & {\color[HTML]{000000} 7.16e-06}                                                              & {\color[HTML]{000000} 0.0104\%}          \\ \hline
{\color[HTML]{000000} \textit{0.0025}}                                                             & {\color[HTML]{000000} 0.00829\%}                                                             & {\color[HTML]{000000} 3.84e-06}                                                              & {\color[HTML]{000000} 0.00840\%}          & {\color[HTML]{000000} \textit{0.0025}}                                                             & {\color[HTML]{000000} 0.0103\%}                                                              & {\color[HTML]{000000} 4.79e-06}                                                              & {\color[HTML]{000000} 0.0103\%}          \\ \hline
{\color[HTML]{000000} \textit{\textbf{0.001}}}                                                     & {\color[HTML]{000000} \textbf{0.00829\%}}                                                    & {\color[HTML]{000000} \textbf{2.51e-06}}                                                     & {\color[HTML]{000000} \textbf{0.00836\%}} & {\color[HTML]{000000} \textit{\textbf{0.001}}}                                                     & {\color[HTML]{000000} \textbf{0.0103\%}}                                                     & {\color[HTML]{000000} \textbf{3.38e-06}}                                                     & {\color[HTML]{000000} \textbf{0.0103\%}} \\ \hline
\end{tabular}%
}
\captionsetup{width=.8\linewidth}
\caption{Test Error and Generalization Gap comparison for different choices of regularization parameters,  No noise in the data.}
\label{tab:PenT0N}
\end{center}
\end{table}

\begin{table}[H]
\begin{center}
\resizebox{\columnwidth}{!}{%
\begin{tabular}{|
>{\columncolor[HTML]{FFFFFF}}c |
>{\columncolor[HTML]{FFFFFF}}c |
>{\columncolor[HTML]{FFFFFF}}c |
>{\columncolor[HTML]{FFFFFF}}c |
>{\columncolor[HTML]{FFFFFF}}c |
>{\columncolor[HTML]{FFFFFF}}c |
>{\columncolor[HTML]{FFFFFF}}c |
>{\columncolor[HTML]{FFFFFF}}c |}
\hline
\multicolumn{4}{|c|}{\cellcolor[HTML]{FFFFFF}{\color[HTML]{000000} \textit{First Component}}}                                                                                                                                                                                                                                               & \multicolumn{4}{c|}{\cellcolor[HTML]{FFFFFF}{\color[HTML]{000000} \textit{Second Component}}}                                                                                                                                                                                                                                               \\ \hline
{\color[HTML]{000000} \textit{\begin{tabular}[c]{@{}c@{}}Regularization\\ Parameter\end{tabular}}} & {\color[HTML]{000000} \textit{\begin{tabular}[c]{@{}c@{}}Baseline\\ Train MSE\end{tabular}}} & {\color[HTML]{000000} \textit{\begin{tabular}[c]{@{}c@{}}Generalization\\ Gap\end{tabular}}} & {\color[HTML]{000000} \textit{Test MSE}} & {\color[HTML]{000000} \textit{\begin{tabular}[c]{@{}c@{}}Regularization\\ Parameter\end{tabular}}} & {\color[HTML]{000000} \textit{\begin{tabular}[c]{@{}c@{}}Baseline\\ Train MSE\end{tabular}}} & {\color[HTML]{000000} \textit{\begin{tabular}[c]{@{}c@{}}Generalization\\ Gap\end{tabular}}} & {\color[HTML]{000000} \textit{Test MSE}} \\ \hline
{\color[HTML]{000000} \textit{0}}                                                                  & {\color[HTML]{000000} 0.0200\%}                                                              & {\color[HTML]{000000} 3.44e-05}                                                              & {\color[HTML]{000000} 0.0210\%}          & {\color[HTML]{000000} \textit{0}}                                                                  & {\color[HTML]{000000} 0.0256\%}                                                              & {\color[HTML]{000000} 1.45e-04}                                                              & {\color[HTML]{000000} 0.0273\%}          \\ \hline
{\color[HTML]{000000} \textit{\textbf{0.01}}}                                                      & {\color[HTML]{000000} \textbf{0.0200\%}}                                                     & {\color[HTML]{000000} \textbf{2.08e-06}}                                                     & {\color[HTML]{000000} \textbf{0.0200\%}} & {\color[HTML]{000000} \textit{0.01}}                                                               & {\color[HTML]{000000} 0.0256\%}                                                              & {\color[HTML]{000000} 5.86e-05}                                                              & {\color[HTML]{000000} 0.0263\%}          \\ \hline
{\color[HTML]{000000} \textit{0.005}}                                                              & {\color[HTML]{000000} 0.0200\%}                                                              & {\color[HTML]{000000} 8.39e-06}                                                              & {\color[HTML]{000000} 0.0202\%}          & {\color[HTML]{000000} \textit{\textbf{0.005}}}                                                     & {\color[HTML]{000000} \textbf{0.0256\%}}                                                     & {\color[HTML]{000000} \textbf{1.31e-05}}                                                     & {\color[HTML]{000000} \textbf{0.0257\%}} \\ \hline
{\color[HTML]{000000} \textit{0.0025}}                                                             & {\color[HTML]{000000} 0.0200\%}                                                              & {\color[HTML]{000000} 3.71e-06}                                                              & {\color[HTML]{000000} 0.0201\%}          & {\color[HTML]{000000} \textit{0.0025}}                                                             & {\color[HTML]{000000} 0.0256\%}                                                              & {\color[HTML]{000000} 9.09e-05}                                                              & {\color[HTML]{000000} 0.0267\%}          \\ \hline
{\color[HTML]{000000} \textit{0.001}}                                                              & {\color[HTML]{000000} 0.0200\%}                                                              & {\color[HTML]{000000} 1.61e-05}                                                              & {\color[HTML]{000000} 0.0204\%}          & {\color[HTML]{000000} \textit{0.001}}                                                              & {\color[HTML]{000000} 0.0256\%}                                                              & {\color[HTML]{000000} 7.44e-05}                                                              & {\color[HTML]{000000} 0.0265\%}          \\ \hline
\end{tabular}%
}
\captionsetup{width=.8\linewidth}
\caption{Test Error and Generalization Gap comparison for different choices of regularization parameters,  $1\%$ noise in the data.}
\label{tab:PenT1N}
\end{center}
\end{table}

\begin{table}[H]
\begin{center}
\resizebox{\columnwidth}{!}{%
\begin{tabular}{|
>{\columncolor[HTML]{FFFFFF}}c |
>{\columncolor[HTML]{FFFFFF}}c |
>{\columncolor[HTML]{FFFFFF}}c |
>{\columncolor[HTML]{FFFFFF}}c |
>{\columncolor[HTML]{FFFFFF}}c |
>{\columncolor[HTML]{FFFFFF}}c |
>{\columncolor[HTML]{FFFFFF}}c |
>{\columncolor[HTML]{FFFFFF}}c |}
\hline
\multicolumn{4}{|c|}{\cellcolor[HTML]{FFFFFF}{\color[HTML]{000000} \textit{First Component}}}                                                                                                                                                                                                                                               & \multicolumn{4}{c|}{\cellcolor[HTML]{FFFFFF}{\color[HTML]{000000} \textit{Second Component}}}                                                                                                                                                                                                                                               \\ \hline
{\color[HTML]{000000} \textit{\begin{tabular}[c]{@{}c@{}}Regularization\\ Parameter\end{tabular}}} & {\color[HTML]{000000} \textit{\begin{tabular}[c]{@{}c@{}}Baseline\\ Train MSE\end{tabular}}} & {\color[HTML]{000000} \textit{\begin{tabular}[c]{@{}c@{}}Generalization\\ Gap\end{tabular}}} & {\color[HTML]{000000} \textit{Test MSE}} & {\color[HTML]{000000} \textit{\begin{tabular}[c]{@{}c@{}}Regularization\\ Parameter\end{tabular}}} & {\color[HTML]{000000} \textit{\begin{tabular}[c]{@{}c@{}}Baseline\\ Train MSE\end{tabular}}} & {\color[HTML]{000000} \textit{\begin{tabular}[c]{@{}c@{}}Generalization\\ Gap\end{tabular}}} & {\color[HTML]{000000} \textit{Test MSE}} \\ \hline
{\color[HTML]{000000} \textit{0}}                                                                  & {\color[HTML]{000000} 0.0609\%}                                                              & {\color[HTML]{000000} 7.47e-05}                                                              & {\color[HTML]{000000} 0.0629\%}          & {\color[HTML]{000000} \textit{0}}                                                                  & {\color[HTML]{000000} 0.0819\%}                                                              & {\color[HTML]{000000} 3.23e-04}                                                              & {\color[HTML]{000000} 0.0854\%}          \\ \hline
{\color[HTML]{000000} \textit{\textbf{0.01}}}                                                      & {\color[HTML]{000000} \textbf{0.0609\%}}                                                     & {\color[HTML]{000000} \textbf{2.82e-05}}                                                     & {\color[HTML]{000000} \textbf{0.0617\%}} & {\color[HTML]{000000} \textit{0.01}}                                                               & {\color[HTML]{000000} 0.0819\%}                                                              & {\color[HTML]{000000} 8.59e-05}                                                              & {\color[HTML]{000000} 0.0828\%}          \\ \hline
{\color[HTML]{000000} \textit{0.005}}                                                              & {\color[HTML]{000000} 0.0609\%}                                                              & {\color[HTML]{000000} 5.04e-05}                                                              & {\color[HTML]{000000} 0.0622\%}          & {\color[HTML]{000000} \textit{\textbf{0.005}}}                                                     & {\color[HTML]{000000} \textbf{0.0819\%}}                                                     & {\color[HTML]{000000} \textbf{6.74e-05}}                                                     & {\color[HTML]{000000} \textbf{0.0826\%}} \\ \hline
{\color[HTML]{000000} \textit{0.0025}}                                                             & {\color[HTML]{000000} 0.0609\%}                                                              & {\color[HTML]{000000} 5.55e-05}                                                              & {\color[HTML]{000000} 0.0624\%}          & {\color[HTML]{000000} \textit{0.0025}}                                                             & {\color[HTML]{000000} 0.0819\%}                                                              & {\color[HTML]{000000} 7.57e-05}                                                              & {\color[HTML]{000000} 0.0827\%}          \\ \hline
{\color[HTML]{000000} \textit{0.001}}                                                              & {\color[HTML]{000000} 0.0609\%}                                                              & {\color[HTML]{000000} 5.24e-05}                                                              & {\color[HTML]{000000} 0.0623\%}          & {\color[HTML]{000000} \textit{0.001}}                                                              & {\color[HTML]{000000} 0.0819\%}                                                              & {\color[HTML]{000000} 9.00e-05}                                                              & {\color[HTML]{000000} 0.0829\%}          \\ \hline
\end{tabular}%
}
\captionsetup{width=.8\linewidth}
\caption{Test Error and Generalization Gap comparison for different choices of regularization parameters,  $2\%$ noise in the data.}
\label{tab:PenT2N}
\end{center}
\end{table}
\noindent
In Tables \ref{tab:PenT0N}, \ref{tab:PenT1N} and \ref{tab:PenT2N} we report the regularization parameter, the baseline train MSE and test MSE and the generalization gap for both components, respectively in the case of noiseless data $1\%$ and $2\%$-noisy data. 
Again the recovery is very good for all networks, with regularized network performing better in all cases. We note again that when more noise is present in the data, a larger regularization parameter is preferable since it gives better test accuracy. We also note that, even if the hyperparameters of the two networks $N_1$ and $N_2$ were the same, the recovery of the first component $f_1$ is always more accurate than the recovery of $f_2$; this is due to the simpler functional form of the function $f_1$.\\ In all cases, however, the test error is  smaller than $0.1\%$.\\
As in the previous two examples, we generated plots of the absolute errors on test data obtained by the non-regularized and best Lipschitz regularized networks. As in previous cases we observed an overall better accuracy obtained by the Lipschitz regularized network for all amounts of noise in the data and largest errors attained at the boundaries of the domain and where the training data was sparse. For brevity we omit these plots here, but they can be found in the additional material section of the paper.

\section{Function Recovery on Non-Trajectory Data}
In this section we report the function reconstruction errors obtained using the regularized and non-regularized networks both with and without noise in the data. As explained in Section 2, since in the numerical examples we used synthetic data, we can compare the reconstruction given by our networks with the true right-hand side of the ODE. In a real-world setting, however, this comparison would not be possible: the only way to evaluate the performance of the network would be by computing the test error as we did in Section 5. We report the function recovery errors and plots only for a 1-dimensional example; similar results were obtained for the other examples proposed in Section 4.
\subsection{Recovery of Non-Autonomous 1D ODE}
\begin{figure}[H]
\centering
\begin{subfigure}{.22\textwidth}
\captionsetup{width=.8\linewidth}
  \centering
  \includegraphics[width=0.99\linewidth, height=0.89\linewidth,valign=t]{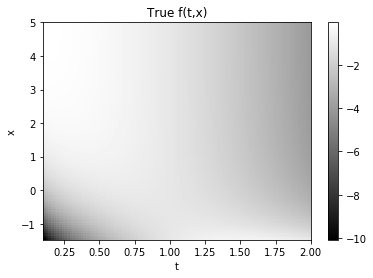}
  \caption*{True Function on full domain. \vspace{36pt}}
  \label{fig:true_e-x_ER}
\end{subfigure}%
\begin{subfigure}{.22\textwidth}
\captionsetup{width=.8\linewidth}
  \centering
  \includegraphics[width=0.95\linewidth, height=0.95\linewidth,valign=t]{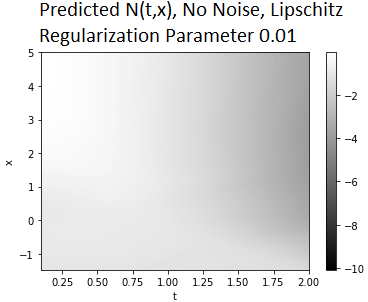}
  \caption*{Recovery on full domain, No Noise in the Data, Lipschitz regularization parameter 0.01}
  \label{fig:0Ne-xER}
\end{subfigure}%
\begin{subfigure}{.22\textwidth}
\captionsetup{width=.8\linewidth}
  \centering
  \includegraphics[width=0.95\linewidth, height=0.95\linewidth,valign=t]{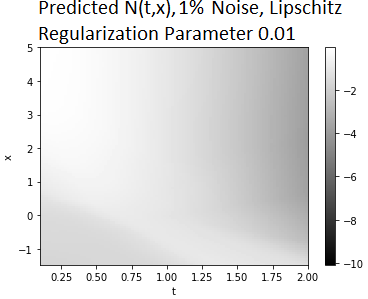}
  \caption*{Recovery on full domain, $1\%$ Noise in the Data, Lipschitz regularization parameter 0.01}
  \label{fig:1Ne-xER}
\end{subfigure}
\begin{subfigure}{.22\textwidth}
\captionsetup{width=.8\linewidth}
  \centering
  \includegraphics[width=0.95\linewidth, height=0.95\linewidth,valign=t]{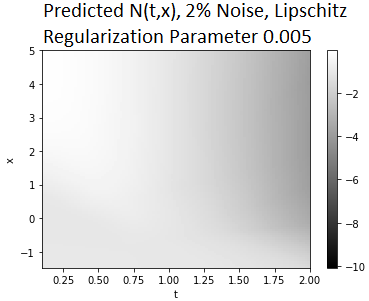}
  \caption*{Recovery on full domain, $2\%$ Noise in the Data, Lipschitz regularization parameter 0.005}
  \label{fig:2Ncosx_ER}
\end{subfigure}
\caption{True function and Recovered functions}
\label{fig:e-x_recovery}
\end{figure}

\begin{figure}[H]
\centering
\begin{subfigure}{.33\textwidth}
\captionsetup{width=.8\linewidth}
  \centering
  \includegraphics[width=0.97\linewidth, height=0.97\linewidth]{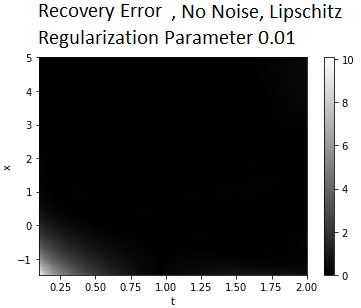}
  \caption*{Recovery error on full domain, No Noise in the Data, Lipschitz regularization parameter 0.01 \vspace{6pt}}
  \label{fig:rec_err_0N}
\end{subfigure}%
\begin{subfigure}{.33\textwidth}
\captionsetup{width=.8\linewidth}
  \centering
  \includegraphics[width=0.95\linewidth,, height=0.95\linewidth]{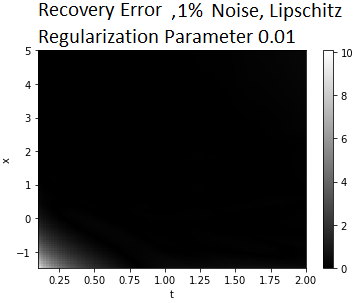}
   \caption*{Recovery error on full domain, $1\%$ Noise in the Data, Lipschitz regularization parameter 0.01}
  \label{fig:rec_err_1N}
\end{subfigure}
\begin{subfigure}{.33\textwidth}
\captionsetup{width=.8\linewidth}
  \centering
  \includegraphics[width=0.95\linewidth,, height=0.95\linewidth]{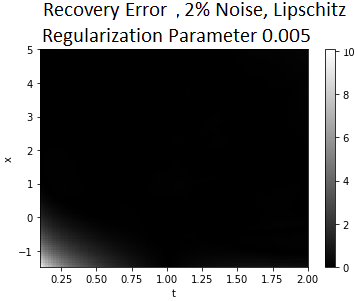}
   \caption*{Recovery error on full domain, $2\%$ Noise in the Data, Lipschitz regularization parameter 0.005}
  \label{fig:rec_err_2N}
\end{subfigure}
\caption{Recovery Error on Full Domain}
\label{fig:rec_err_e-x}
\end{figure}

We report here the recovery errors on the full domain for the ODE: $\dot{x}(t) = e^{-x}\log(t) -t^2$. The spatio-temporal domain in which we recover the function $e^{-x}\log(t) -t^2$ is composed of couples $(t,x) \in [0.1,2] \times [-1.5,5]$ where the original data was generated. We see that the recovery error is worse than the test error obtained in the previous section because of the bad sampling of the function domain. 
\begin{table}[H]
\resizebox{\columnwidth}{!}{%
\parbox{.45\linewidth}{
\centering
\begin{tabular}{|
>{\columncolor[HTML]{FFFFFF}}c |
>{\columncolor[HTML]{FFFFFF}}c |}
\hline
{\color[HTML]{000000} \textit{\begin{tabular}[c]{@{}c@{}}Regularization \\ parameter\end{tabular}}} & {\color[HTML]{000000} \textit{\begin{tabular}[c]{@{}c@{}}Recovery Error \\ on Full Domain\end{tabular}}} \\ \hline
{\color[HTML]{000000} \textit{0}}                                                                   & {\color[HTML]{000000} 1.82\%}                                                                            \\ \hline
{\color[HTML]{000000} \textit{\textbf{0.01}}}                                                       & {\color[HTML]{000000} \textbf{1.79\%}}                                                                   \\ \hline
{\color[HTML]{000000} \textit{0.005}}                                                               & {\color[HTML]{000000} 1.90\%}                                                                            \\ \hline
{\color[HTML]{000000} \textit{0.0025}}                                                              & {\color[HTML]{000000} 1.83\%}                                                                            \\ \hline
{\color[HTML]{000000} \textit{0.001}}                                                               & {\color[HTML]{000000} 1.84\%}                                                                            \\ \hline
\end{tabular}
\captionsetup{width=.8\linewidth}
\caption*{Recovery Error, No Noise in the Data}%
}
\parbox{.45\linewidth}{
\centering
\begin{tabular}{|
>{\columncolor[HTML]{FFFFFF}}c |
>{\columncolor[HTML]{FFFFFF}}c |}
\hline
{\color[HTML]{000000} \textit{\begin{tabular}[c]{@{}c@{}}Regularization \\ parameter\end{tabular}}} & {\color[HTML]{000000} \textit{\begin{tabular}[c]{@{}c@{}}Recovery Error \\ on Full Domain\end{tabular}}} \\ \hline
{\color[HTML]{000000} \textit{0}}                                                                   & {\color[HTML]{000000} 1.76\%}                                                                            \\ \hline
{\color[HTML]{000000} \textit{\textbf{0.01}}}                                                       & {\color[HTML]{000000} \textbf{1.30\%}}                                                                   \\ \hline
{\color[HTML]{000000} \textit{0.005}}                                                               & {\color[HTML]{000000} 1.80\%}                                                                            \\ \hline
{\color[HTML]{000000} \textit{0.0025}}                                                              & {\color[HTML]{000000} 1.78\%}                                                                            \\ \hline
{\color[HTML]{000000} \textit{0.001}}                                                               & {\color[HTML]{000000} 1.64\%}                                                                            \\ \hline
\end{tabular}
\captionsetup{width=.8\linewidth}
\caption*{Recovery Error, $1\%$ Noise in the Data}%
}
\parbox{.45\linewidth}{
\centering
\begin{tabular}{|
>{\columncolor[HTML]{FFFFFF}}c |
>{\columncolor[HTML]{FFFFFF}}c |}
\hline
{\color[HTML]{000000} \textit{\begin{tabular}[c]{@{}c@{}}Regularization \\ parameter\end{tabular}}} & {\color[HTML]{000000} \textit{\begin{tabular}[c]{@{}c@{}}Recovery Error \\ on Full Domain\end{tabular}}} \\ \hline
{\color[HTML]{000000} \textit{0}}                                                                   & {\color[HTML]{000000} 1.71\%}                                                                            \\ \hline
{\color[HTML]{000000} \textit{0.01}}                                                                & {\color[HTML]{000000} 1.72\%}                                                                            \\ \hline
{\color[HTML]{000000} \textit{\textbf{0.005}}}                                                      & {\color[HTML]{000000} \textbf{1.65\%}}                                                                   \\ \hline
{\color[HTML]{000000} \textit{0.0025}}                                                              & {\color[HTML]{000000} 1.83\%}                                                                            \\ \hline
{\color[HTML]{000000} \textit{0.001}}                                                               & {\color[HTML]{000000} 1.70\%}                                                                            \\ \hline
\end{tabular}
\captionsetup{width=.8\linewidth}
\caption*{Recovery Error, $2\%$ Noise in the Data}%
}}
\caption{Recovery Error on Full Domain}
\label{Rece-x}
\end{table}
\noindent
In Table \ref{Rece-x} we report the relative mean recovery error on the full domain. The error is computed as the mean relative difference between $f(t,x)$ and $N(t,x)$ on couples $(t,x)$ on a regular $100\times 100$ grid in the $tx-$domain.\\
We see that, even if the regularization does not improve the recovery error for all choices of the regularization parameter, the best accuracy is obtained in all cases by a regularized network, showing that regularized network are able to better generalize from part of the domain where data was available to parts of the domain where the data was not present.
\medskip\\
Finally, below are the plots of the true function and of the recovery obtained by the trained networks which attain lowest recovery errors in the $tx-$domain; the gray level represent the value of the function for each couple $(t,x)$ (see Figure \ref{fig:e-x_recovery}). We note that where less data was present initially, the recovery of the function is less accurate, both in the non-regularized and regularized cases. We also provide plots of the recovery error in the three cases; the gray level represents the magnitude of the error (see Figure \ref{fig:rec_err_e-x}). From the error plots we can clearly see that the maximum error is attained in the lower left corner of the domain. In this part of the domain not only no training data was available, but also the function $f(t,x)$ changes suddenly from a value of around $-2$ to a value of $-10$. It is expected then that the network is not able to detect this sudden change in the function values since this was not present in the training data. We notice also that, even if in the upper right corner no training data was available, the networks are able to extrapolate well the function behaviour there. This is because the true function $f(t,x)$ attains similar values in the upper right corner as in the data-dense part of the domain (the central part of the domain). 

\begin{remark}
We explicitly want to notice that the error obtained in the recovery is strongly influenced by the quality of the target data. Specifically, in presence of noise if the spline approximation of the noisy solution is not accurate, the target data may not be representative of the derivative function. As a result, the testing error may be significantly smaller than the recovery error on the full domain, not only because the full domain is not well sampled, but also because the quality of the data was not very good. This also explains why the best regularization parameters obtained when comparing test errors are different than the ones obtained when comparing recovery errors. This shows the importance of a good preprocessing of the data in case of real-world data and the importance of further study in this direction.
\end{remark}

% \begin{remark}
% In contrast with what we expected, a higher amount of noise in the data does not result in the best regularization parameter being larger. We think this behaviour is due to the preprocessing of our data using splines. When using raw noisy trajectories, a larger noise results in very large and inaccurate point-wise derivatives. For this reason, we would expect that the higher the noise amount, the larger constraint on the Lipschitz constant of the network is needed to get a good test accuracy. However, when preprocessing the data using splines, the smoothing of the trajectories results already in smaller and more regular point-wise derivatives, making the magnitude of the regularization parameter less important (but as seen in the examples, not useless).
% \end{remark}
\section{Conclusion}
In this paper we use neural network to learn governing equations from data. Differently than other papers that use neural networks for system identification, we add a Lipschitz regularization term to our loss function to force the Lipschitz constant of the network to be small. This regularization results in a smoother approximating function and better generalization properties when compared with non-regularized models, especially in presence of noise. These results are in line with the theoretical work of Calder and Oberman \cite{oberman2018lipschitz} in which they prove that networks with Lipschitz regularization converge and generalize. The results shown in the examples, which are representative of a larger testing activity with several different types of right-hand sides $f(x,t)$, show another advantage of our method: since neural networks are universal approximators, we do not need any prior knowledge on the ODE system, in contrast with sparse regression approaches in which a library of candidate functions has to be defined. Finally, since our method is applied componentwise, it can be used to identify systems of any dimension, which makes it a valuable approach for high-dimensional real-world problems.
\medskip\\
We propose examples for synthetic data for autonomous and non autonomous equations of different dimensions and in all cases we observe a better test error on trajectory data when using the Lipschitz regularization term in the loss function. Since we use synthetic data we also have access to the true right-hand side function $f$. For this reason, we also record recovery errors on non-trajectory data by comparing the output of network $N$ with the true function $f(t,x)$ on arbitrary couples $(t,x)$ in the domain of $f$. Again the best accuracy is obtained by regularized networks, showing again that the Lipschitz regularized networks are able to generalize better to unseen data than non-regularized ones. Our experiments also show that accurate interpolation of measurement data is necessary to obtain reliable derivative approximation. Since these are used as target data, their quality strongly influences the results obtained especially on non-trajectory data; better preprocessing techniques will be object of future work.
\medskip\\
In the future we would like to apply our method to real-world data, for which we expect the data preprocessing and denoising to be central in order to obtain accurate results.\\
We would also like to generalize this method to learn partial differential equations.\\
Finally, in contrast with the most common choices of regularization terms found in machine learning literature, in this work we impose a regularization on the network statistical geometric mapping properties, instead of on its parameters. Of course, since the minimization of the loss function is done with respect to the network parameters, the Lipschitz regularization term results in an implicit constraint on such parameters. In the future we would like to study theoretically this regularization term, how it is related to the size of the weights of the network, in line with Bartlett work about generalization \cite{bartlett1997valid}, and explicitly express it as a constraint on the network parameters. This latter study would result in a much faster model since obtaining a good approximation of the Lipschitz constant of the network can be computationally very expensive.

\section{Acknowledgements}
Luca Capogna is partially supported by NSF DMS 1955992 and Simons Collaboration Grant for Mathematicians 585688.\\
Giovanna Citti is partially supported by the EU Horizon 2020 project GHAIA,  MCSA RISE project GA No 777822.\\
Results in this paper were obtained in part using a high-performance computing system acquired through NSF MRI grant DMS-1337943 to WPI.\\
The authors would like to thank Randy Paffenroth for his insightful
advice.

\newpage
\bibliographystyle{plain}
\small
\bibliography{bibib}

\begin{thebibliography}{10}

\bibitem{abu2012learning}
Yaser~S Abu-Mostafa, Malik Magdon-Ismail, and Hsuan-Tien Lin.
\newblock {\em Learning from data}, volume~4.
\newblock AMLBook New York, NY, USA:, 2012.

\bibitem{bartlett1997valid}
Peter~L Bartlett.
\newblock For valid generalization the size of the weights is more important
  than the size of the network.
\newblock In {\em Advances in neural information processing systems}, pages
  134--140, 1997.

\bibitem{berg2017neural}
Jens Berg and Kaj Nystr{\"o}m.
\newblock Neural network augmented inverse problems for pdes.
\newblock {\em arXiv preprint arXiv:1712.09685}, 2017.

\bibitem{berg2019data}
Jens Berg and Kaj Nystr{\"o}m.
\newblock Data-driven discovery of pdes in complex datasets.
\newblock {\em Journal of Computational Physics}, 384:239--252, 2019.

\bibitem{billings2013nonlinear}
Stephen~A Billings.
\newblock {\em Nonlinear system identification: NARMAX methods in the time,
  frequency, and spatio-temporal domains}.
\newblock John Wiley \& Sons, 2013.

\bibitem{brunton2016discovering}
Steven~L Brunton, Joshua~L Proctor, and J~Nathan Kutz.
\newblock Discovering governing equations from data by sparse identification of
  nonlinear dynamical systems.
\newblock {\em Proceedings of the national academy of sciences},
  113(15):3932--3937, 2016.

\bibitem{calliess2015bayesian}
Jan-Peter Calliess.
\newblock Bayesian lipschitz constant estimation and quadrature.
\newblock {\em Advances in Neural Information Processing Systems}, 2015.

\bibitem{dinh2010dynamic}
Huyen Dinh, Shubhendu Bhasin, and Warren~E Dixon.
\newblock Dynamic neural network-based robust identification and control of a
  class of nonlinear systems.
\newblock In {\em 49th IEEE Conference on Decision and Control (CDC)}, pages
  5536--5541. IEEE, 2010.

\bibitem{garcia2015data}
Salvador Garc{\'\i}a, Juli{\'a}n Luengo, and Francisco Herrera.
\newblock {\em Data preprocessing in data mining}.
\newblock Springer, 2015.

\bibitem{gouk2018regularisation}
Henry Gouk, Eibe Frank, Bernhard Pfahringer, and Michael Cree.
\newblock Regularisation of neural networks by enforcing lipschitz continuity.
\newblock {\em arXiv preprint arXiv:1804.04368}, 2018.

\bibitem{hasan2020learning}
Ali Hasan, Jo{\~a}o~M Pereira, Robert Ravier, Sina Farsiu, and Vahid Tarokh.
\newblock Learning partial differential equations from data using neural
  networks.
\newblock In {\em ICASSP 2020-2020 IEEE International Conference on Acoustics,
  Speech and Signal Processing (ICASSP)}, pages 3962--3966. IEEE, 2020.

\bibitem{kotsiantis2006data}
SB~Kotsiantis, Dimitris Kanellopoulos, and PE~Pintelas.
\newblock Data preprocessing for supervised leaning.
\newblock {\em International Journal of Computer Science}, 1(2):111--117, 2006.

\bibitem{kuschewski1993application}
John~G Kuschewski, Stefen Hui, and Stanislaw~H Zak.
\newblock Application of feedforward neural networks to dynamical system
  identification and control.
\newblock {\em IEEE Transactions on Control Systems Technology}, 1(1):37--49,
  1993.

\bibitem{leshno1993multilayer}
Moshe Leshno, Vladimir~Ya Lin, Allan Pinkus, and Shimon Schocken.
\newblock Multilayer feedforward networks with a nonpolynomial activation
  function can approximate any function.
\newblock {\em Neural networks}, 6(6):861--867, 1993.

\bibitem{ljung2010perspectives}
Lennart Ljung.
\newblock Perspectives on system identification.
\newblock {\em Annual Reviews in Control}, 34(1):1--12, 2010.

\bibitem{narendra1992neural}
Kumpati~S Narendra and Kannan Parthasarathy.
\newblock Neural networks and dynamical systems.
\newblock {\em International Journal of Approximate Reasoning}, 6(2):109--131,
  1992.

\bibitem{nelles2013nonlinear}
Oliver Nelles.
\newblock {\em Nonlinear system identification: from classical approaches to
  neural networks and fuzzy models}.
\newblock Springer Science \& Business Media, 2013.

\bibitem{oberman2018lipschitz}
Adam~M Oberman and Jeff Calder.
\newblock Lipschitz regularized deep neural networks converge and generalize.
\newblock {\em arXiv preprint arXiv:1808.09540}, 2018.

\bibitem{ogunmolu2016nonlinear}
Olalekan Ogunmolu, Xuejun Gu, Steve Jiang, and Nicholas Gans.
\newblock Nonlinear systems identification using deep dynamic neural networks.
\newblock {\em arXiv preprint arXiv:1610.01439}, 2016.

\bibitem{pauli2020training}
Patricia Pauli, Anne Koch, Julian Berberich, and Frank Allg{\"o}wer.
\newblock Training robust neural networks using lipschitz bounds.
\newblock {\em arXiv preprint arXiv:2005.02929}, 2020.

\bibitem{qin2019data}
Tong Qin, Kailiang Wu, and Dongbin Xiu.
\newblock Data driven governing equations approximation using deep neural
  networks.
\newblock {\em Journal of Computational Physics}, 395:620--635, 2019.

\bibitem{raissi2018deep}
Maziar Raissi.
\newblock Deep hidden physics models: Deep learning of nonlinear partial
  differential equations.
\newblock {\em The Journal of Machine Learning Research}, 19(1):932--955, 2018.

\bibitem{raissi2018hidden}
Maziar Raissi and George~Em Karniadakis.
\newblock Hidden physics models: Machine learning of nonlinear partial
  differential equations.
\newblock {\em Journal of Computational Physics}, 357:125--141, 2018.

\bibitem{raissi2017machine}
Maziar Raissi, Paris Perdikaris, and George~Em Karniadakis.
\newblock Machine learning of linear differential equations using gaussian
  processes.
\newblock {\em Journal of Computational Physics}, 348:683--693, 2017.

\bibitem{raissi2018numerical}
Maziar Raissi, Paris Perdikaris, and George~Em Karniadakis.
\newblock Numerical gaussian processes for time-dependent and nonlinear partial
  differential equations.
\newblock {\em SIAM Journal on Scientific Computing}, 40(1):A172--A198, 2018.

\bibitem{rudy2017data}
Samuel~H Rudy, Steven~L Brunton, Joshua~L Proctor, and J~Nathan Kutz.
\newblock Data-driven discovery of partial differential equations.
\newblock {\em Science Advances}, 3(4):e1602614, 2017.

\bibitem{sahoo2018learning}
Subham~S Sahoo, Christoph~H Lampert, and Georg Martius.
\newblock Learning equations for extrapolation and control.
\newblock {\em arXiv preprint arXiv:1806.07259}, 2018.

\bibitem{schaeffer2017learning}
Hayden Schaeffer.
\newblock Learning partial differential equations via data discovery and sparse
  optimization.
\newblock {\em Proceedings of the Royal Society A: Mathematical, Physical and
  Engineering Sciences}, 473(2197):20160446, 2017.

\bibitem{schaeffer2013sparse}
Hayden Schaeffer, Russel Caflisch, Cory~D Hauck, and Stanley Osher.
\newblock Sparse dynamics for partial differential equations.
\newblock {\em Proceedings of the National Academy of Sciences},
  110(17):6634--6639, 2013.

\bibitem{wang2006fully}
Jeen-Shing Wang and Yen-Ping Chen.
\newblock A fully automated recurrent neural network for unknown dynamic system
  identification and control.
\newblock {\em IEEE Transactions on Circuits and Systems I: Regular Papers},
  53(6):1363--1372, 2006.

\bibitem{xu2012robustness}
Huan Xu and Shie Mannor.
\newblock Robustness and generalization.
\newblock {\em Machine learning}, 86(3):391--423, 2012.

\end{thebibliography}

\appendix
\section{Supplementary Material}
\subsection{Error Plots for Lotka Volterra System}
\begin{figure}[H]
\centering
\begin{subfigure}{.5\textwidth}
\captionsetup{width=.8\linewidth}
  \centering
  \includegraphics[width=0.85\linewidth,height=0.9\linewidth]{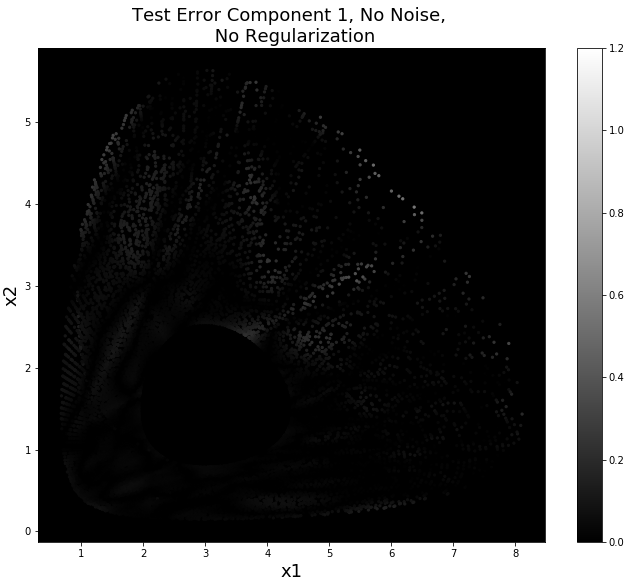}
  \caption{Test error non-regularized network, No noise in the data, first component. \vspace{12pt}}
\end{subfigure}%
\begin{subfigure}{.5\textwidth}
\captionsetup{width=.8\linewidth}
  \centering
  \includegraphics[width=0.85\linewidth,height=0.9\linewidth]{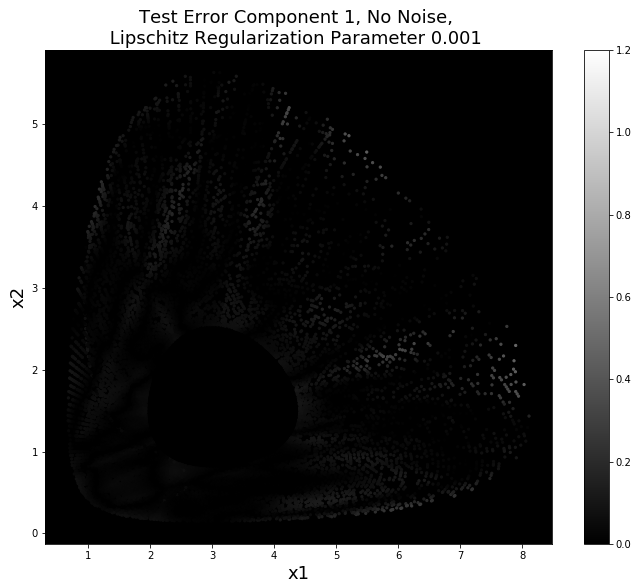}
  \caption{Test error Lipschitz regularized network with parameter 0.001, no noise in the data.}
\end{subfigure}
\begin{subfigure}{.5\textwidth}
\captionsetup{width=.8\linewidth}
  \centering
  \includegraphics[width=0.85\linewidth,height=0.9\linewidth]{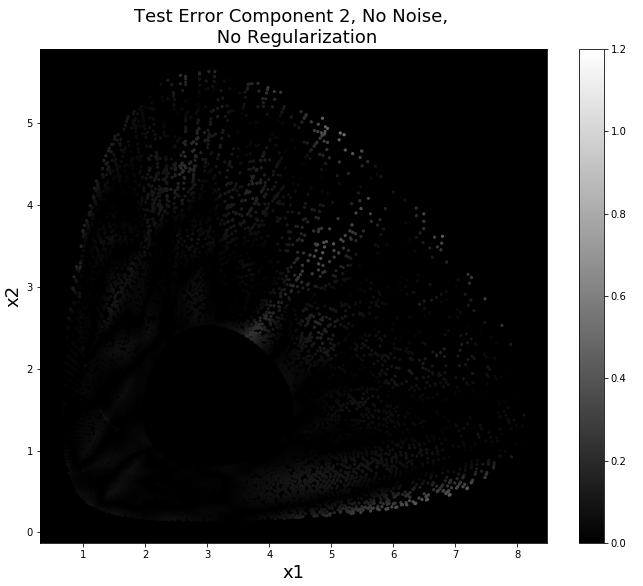}
  \caption{Test error non-regularized network, no noise in the data, second component. \\}
\end{subfigure}%
\begin{subfigure}{.5\textwidth}
\captionsetup{width=.8\linewidth}
  \centering
  \includegraphics[width=0.85\linewidth, height=0.9\linewidth]{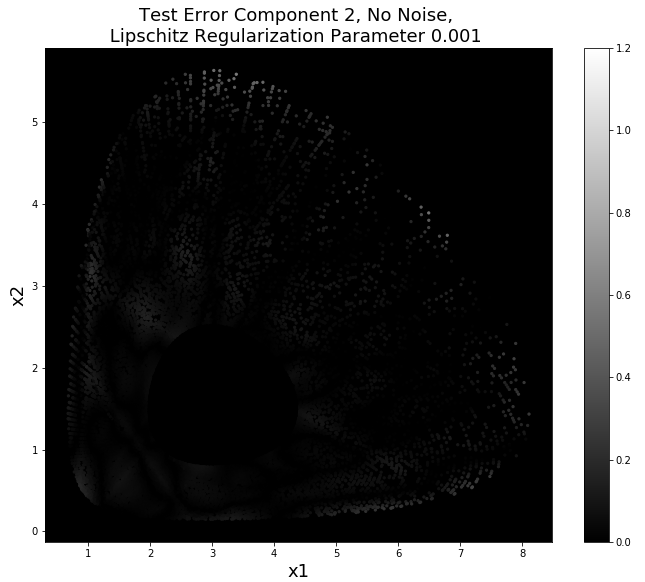}
  \caption{Test error Lipschitz regularized network with parameter 0.001, no noise data, second component.}
\end{subfigure}
\caption{Test error comparison, no noise in the data.}
\end{figure}

\begin{figure}[H]
\centering
\begin{subfigure}{.5\textwidth}
\captionsetup{width=.8\linewidth}
  \centering
  \includegraphics[width=0.85\linewidth,height=0.9\linewidth]{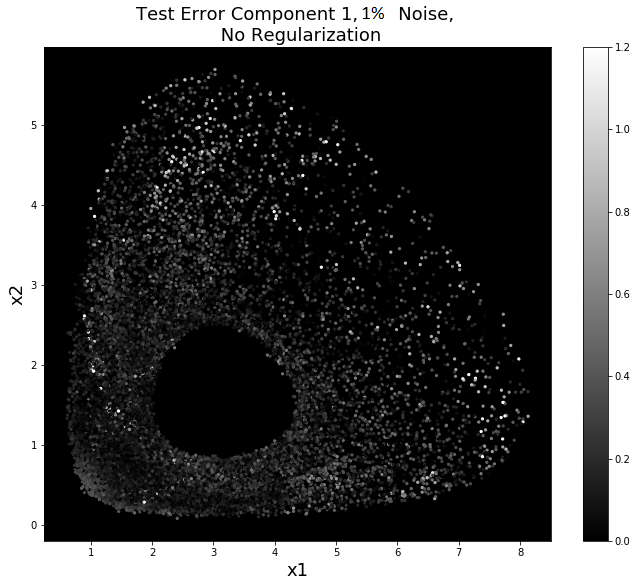}
  \caption{Test error non-regularized network, $1\%$ noise in the data, first component. \\}
\end{subfigure}%
\begin{subfigure}{.5\textwidth}
\captionsetup{width=.8\linewidth}
  \centering
  \includegraphics[width=0.85\linewidth,height=0.9\linewidth]{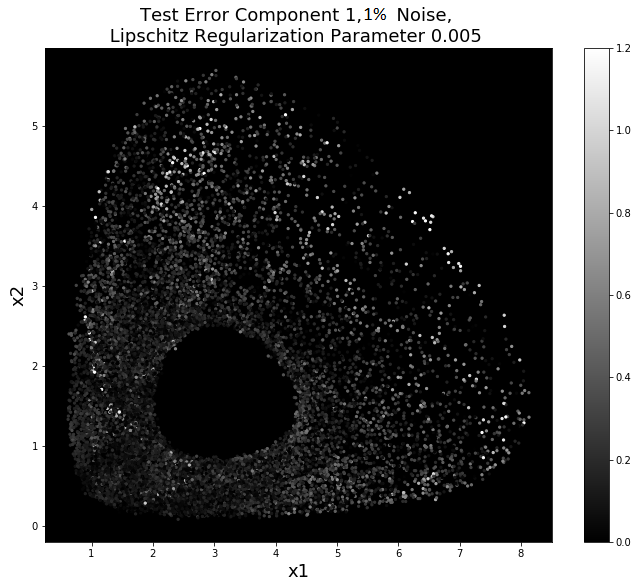}
  \caption{Test error Lipschitz regularized network with parameter 0.005, $1\%$ noise in the data, first component.}
\end{subfigure}
\begin{subfigure}{.5\textwidth}
\captionsetup{width=.8\linewidth}
  \centering
  \includegraphics[width=0.85\linewidth,height=0.9\linewidth]{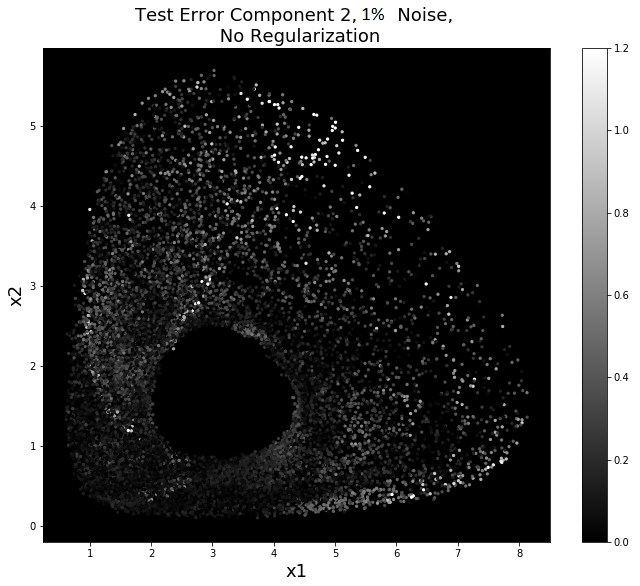}
  \caption{Test error non-regularized network, $1\%$ noise in the data, second component.}
\end{subfigure}%
\begin{subfigure}{.5\textwidth}
\captionsetup{width=.8\linewidth}
  \centering
  \includegraphics[width=0.85\linewidth, height=0.9\linewidth]{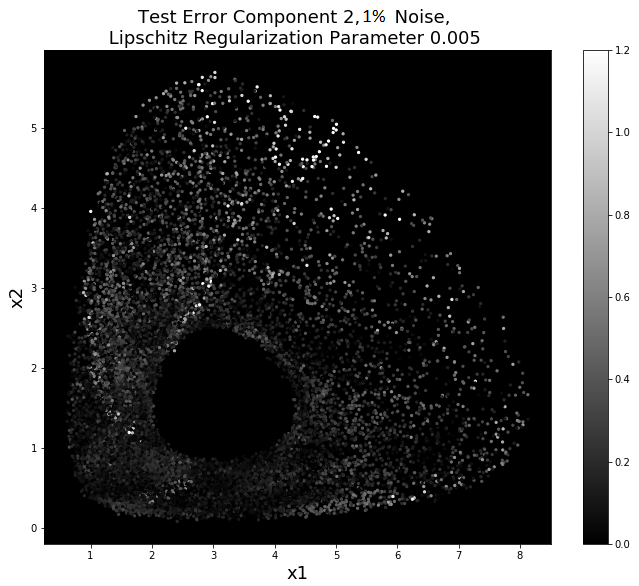}
  \caption{Test error Lipschitz regularized network with parameter 0.005, $1\%$ noise data, second component.}
\end{subfigure}
\caption{Test error comparison, $1\%$ noise in the data.}
\end{figure}

\begin{figure}[H]
\centering
\begin{subfigure}{.5\textwidth}
\captionsetup{width=.8\linewidth}
  \centering
  \includegraphics[width=0.85\linewidth,height=0.9\linewidth]{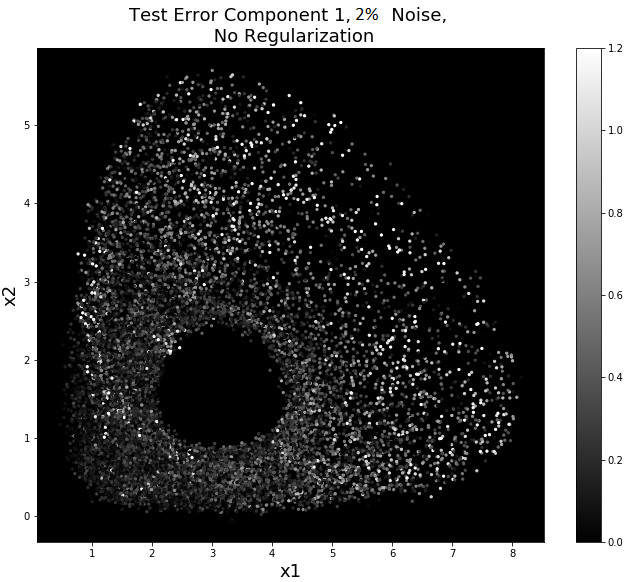}
  \caption{Test error non-regularized network, $2\%$ noise in the data, first component. \\}
\end{subfigure}%
\begin{subfigure}{.5\textwidth}
\captionsetup{width=.8\linewidth}
  \centering
  \includegraphics[width=0.85\linewidth,height=0.9\linewidth]{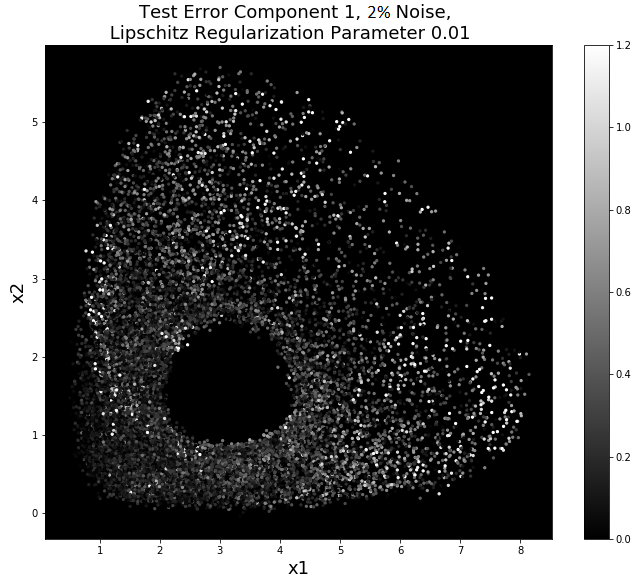}
  \caption{Test error Lipschitz regularized network with parameter 0.01, $2\%$ noise in the data, first component.}
  \label{fig:E2LVLR1}
\end{subfigure}
\begin{subfigure}{.5\textwidth}
\captionsetup{width=.8\linewidth}
  \centering
  \includegraphics[width=0.85\linewidth,height=0.9\linewidth]{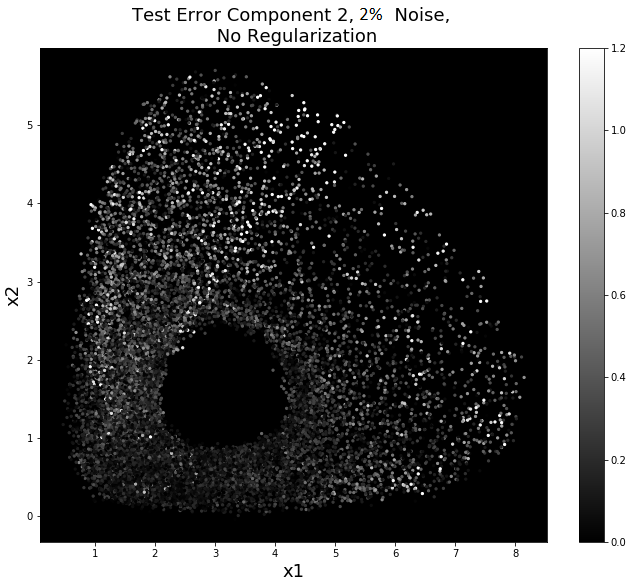}
  \caption{Test error non-regularized network, $2\%$ noise in the data, second component.}
\end{subfigure}%
\begin{subfigure}{.5\textwidth}
\captionsetup{width=.8\linewidth}
  \centering
  \includegraphics[width=0.85\linewidth, height=0.9\linewidth]{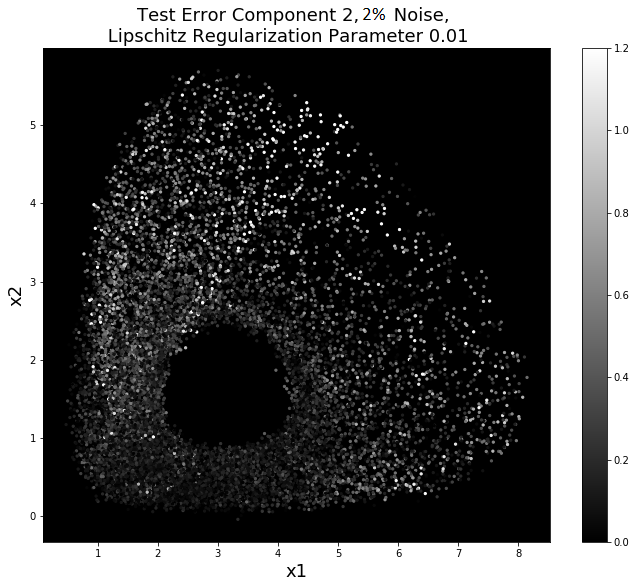}
  \caption{Test error Lipschitz regularized network with parameter 0.01, $2\%$ noise data, second component.}
\end{subfigure}
\caption{Test error comparison, $2\%$ noise in the data.}
\end{figure}
\subsection{Error Plots for Second Order Non-Autonomous ODE}
\begin{figure}[H]
\centering
\begin{subfigure}{.5\textwidth}
\captionsetup{width=.8\linewidth}
  \centering
  \includegraphics[width=0.85\linewidth,height=0.9\linewidth]{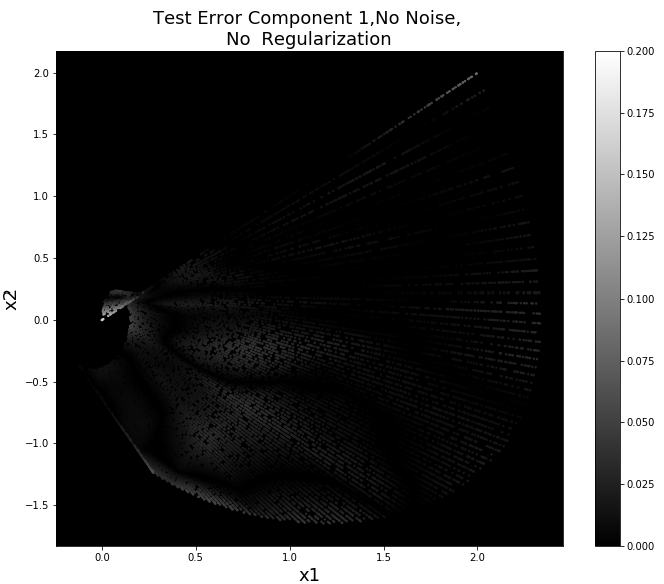}
  \caption{Test error non-regularized network, No noise in the data, first component. \\}
\end{subfigure}%
\begin{subfigure}{.5\textwidth}
\captionsetup{width=.8\linewidth}
  \centering
  \includegraphics[width=0.85\linewidth,height=0.9\linewidth]{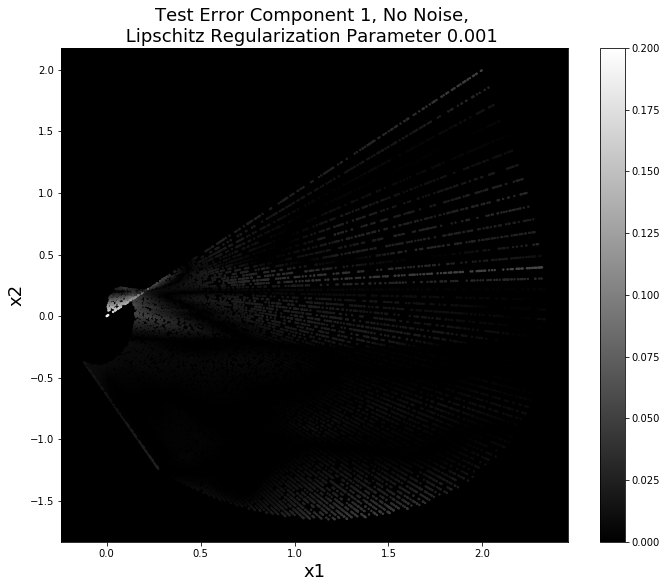}
  \caption{Test error Lipschitz regularized network with parameter 0.001, no noise in the data.}
\end{subfigure}
\begin{subfigure}{.5\textwidth}
\captionsetup{width=.8\linewidth}
  \centering
  \includegraphics[width=0.85\linewidth,height=0.9\linewidth]{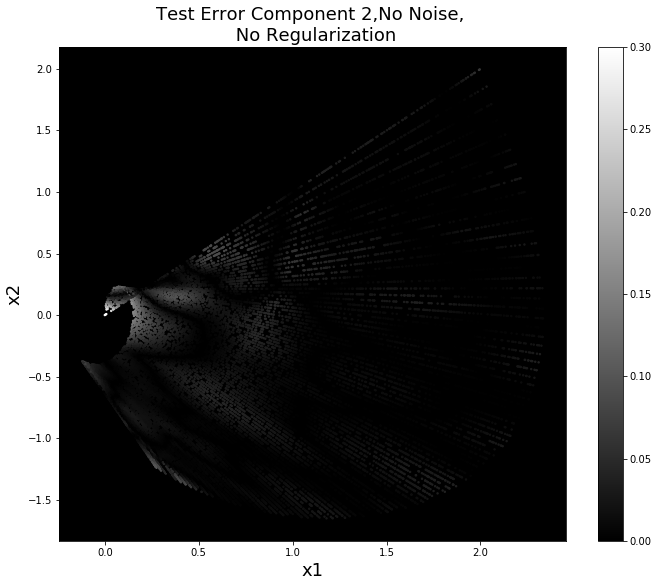}
  \caption{Test error non-regularized network, no noise in the data, second component. \\}
\end{subfigure}%
\begin{subfigure}{.5\textwidth}
\captionsetup{width=.8\linewidth}
  \centering
  \includegraphics[width=0.85\linewidth, height=0.9\linewidth]{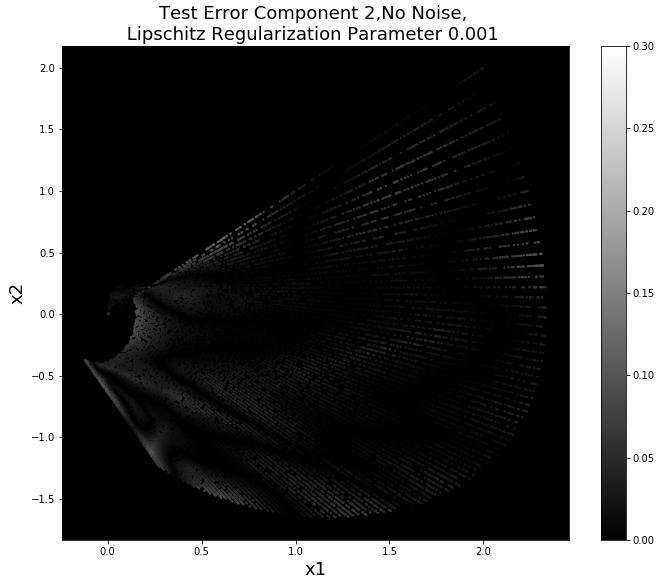}
  \caption{Test error Lipschitz regularized network with parameter 0.01, no noise data, second component.}
\end{subfigure}
\caption{Test error comparison, no noise in the data, first component.}
\end{figure}

\begin{figure}[H]
\centering
\begin{subfigure}{.5\textwidth}
\captionsetup{width=.8\linewidth}
  \centering
  \includegraphics[width=0.85\linewidth,height=0.9\linewidth]{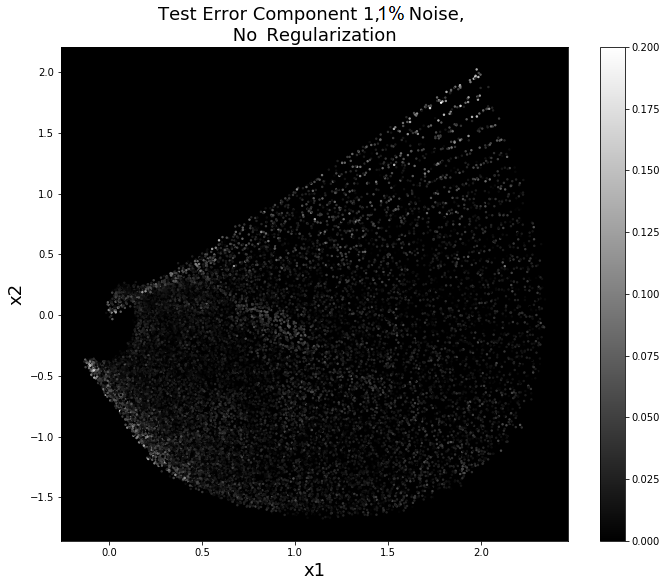}
  \caption{Test error non-regularized network, $1\%$ noise in the data, first component. \\}
\end{subfigure}%
\begin{subfigure}{.5\textwidth}
\captionsetup{width=.8\linewidth}
  \centering
  \includegraphics[width=0.85\linewidth,height=0.9\linewidth]{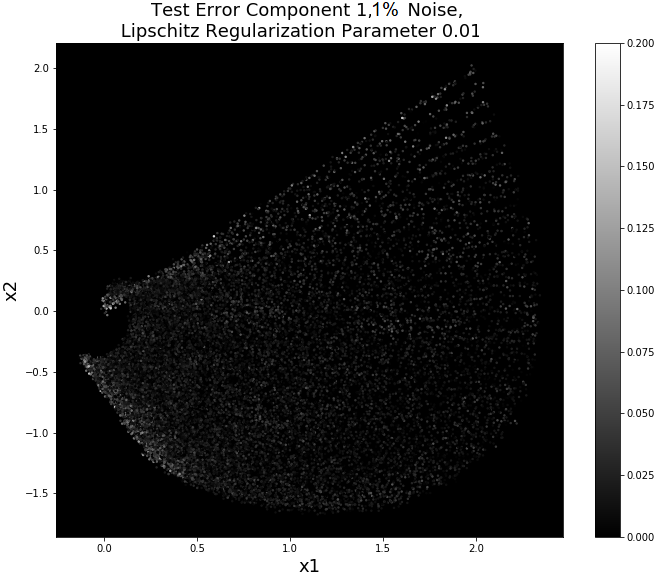}
  \caption{Test error Lipschitz regularized network with parameter 0.01, $1\%$ noise in the data, first component.}
\end{subfigure}
\begin{subfigure}{.5\textwidth}
\captionsetup{width=.8\linewidth}
  \centering
  \includegraphics[width=0.85\linewidth,height=0.9\linewidth]{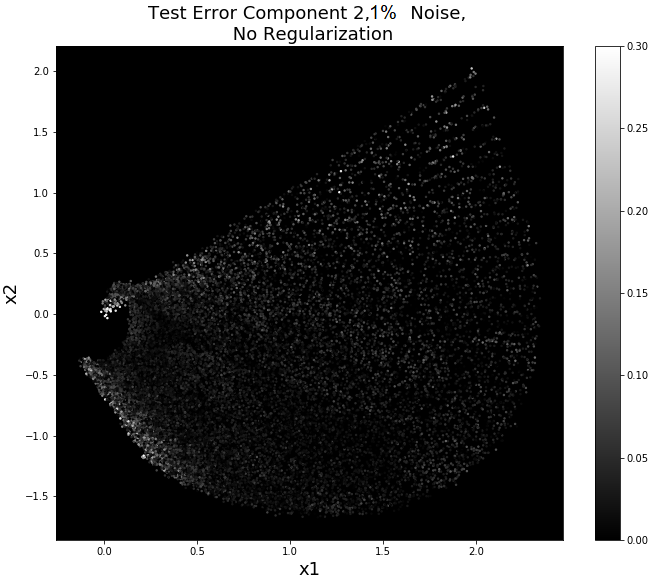}
  \caption{Test error non-regularized network, $1\%$ noise in the data, second component.}
\end{subfigure}%
\begin{subfigure}{.5\textwidth}
\captionsetup{width=.8\linewidth}
  \centering
  \includegraphics[width=0.85\linewidth, height=0.9\linewidth]{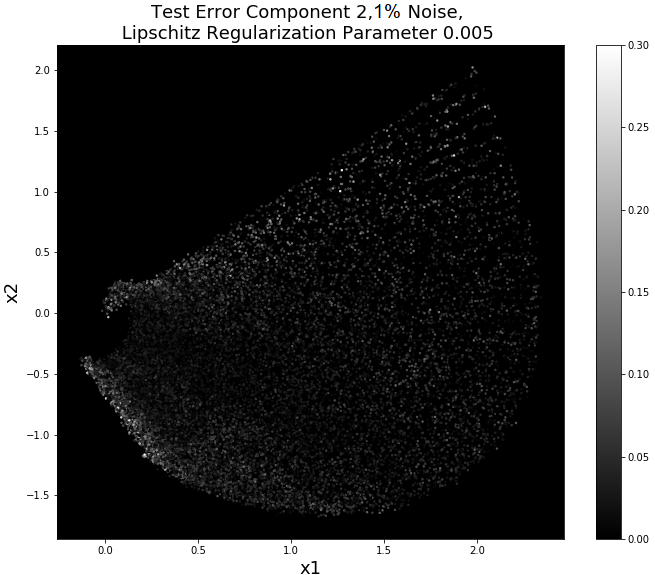}
  \caption{Test error Lipschitz regularized network with parameter 0.005, $1\%$ noise data, second component.}
\end{subfigure}
\caption{Test error comparison, $1\%$ noise in the data.}
\label{fig:Err1NPend1}
\end{figure}

\begin{figure}[H]
\centering
\begin{subfigure}{.5\textwidth}
\captionsetup{width=.8\linewidth}
  \centering
  \includegraphics[width=0.85\linewidth,height=0.9\linewidth]{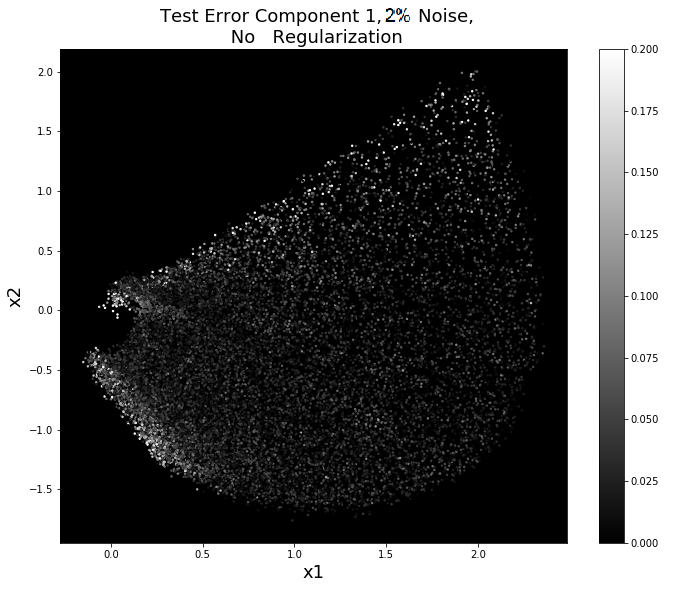}
  \caption{Test error non-regularized network, $2\%$ noise in the data, first component. \\}
\end{subfigure}%
\begin{subfigure}{.5\textwidth}
\captionsetup{width=.8\linewidth}
  \centering
  \includegraphics[width=0.85\linewidth,height=0.9\linewidth]{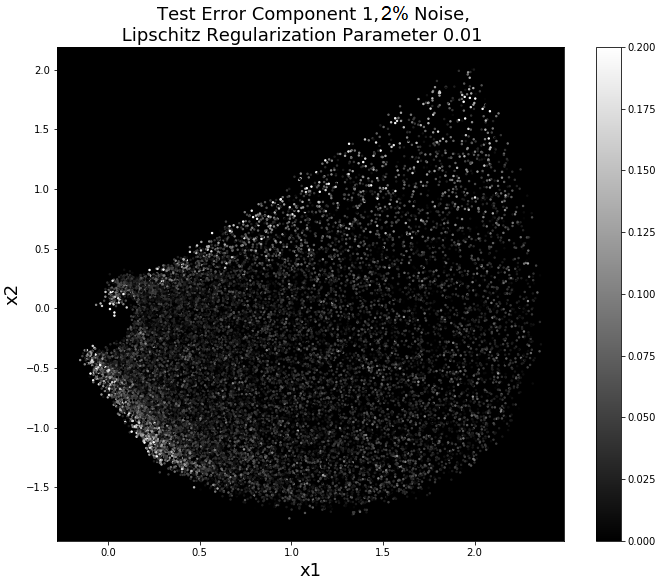}
  \caption{Test error Lipschitz regularized network with parameter 0.01, $2\%$ noise in the data, first component.}
\end{subfigure}
\begin{subfigure}{.5\textwidth}
\captionsetup{width=.8\linewidth}
  \centering
  \includegraphics[width=0.85\linewidth,height=0.9\linewidth]{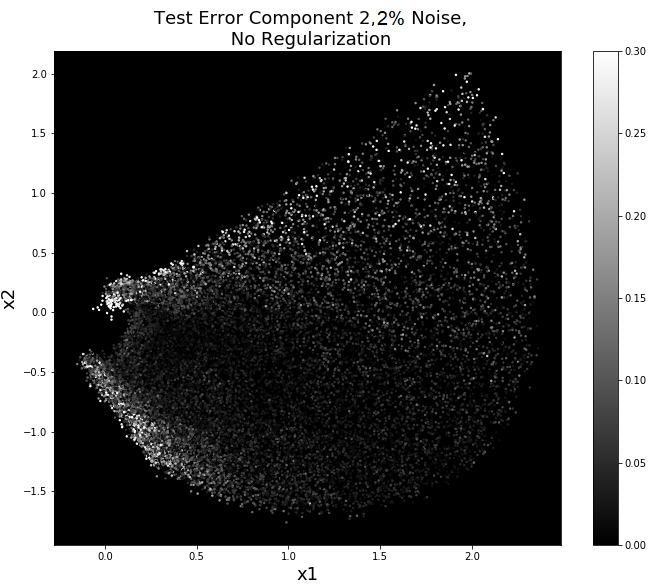}
  \caption{Test error non-regularized network, $2\%$ noise in the data, second component.}
\end{subfigure}%
\begin{subfigure}{.5\textwidth}
\captionsetup{width=.8\linewidth}
  \centering
  \includegraphics[width=0.85\linewidth, height=0.9\linewidth]{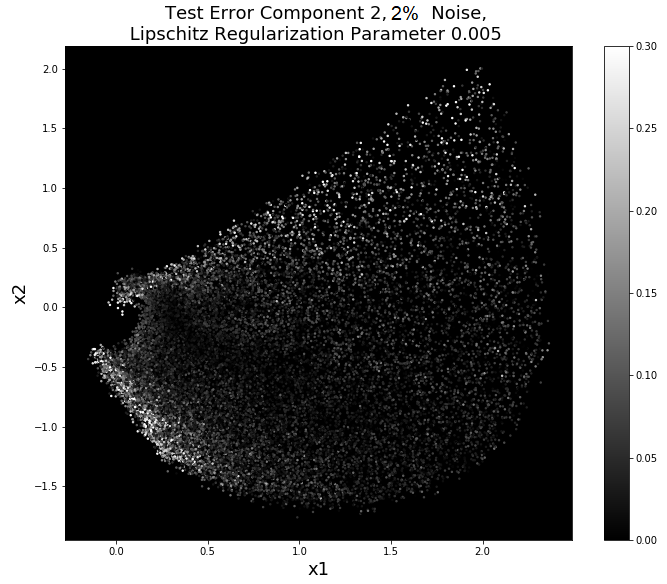}
  \caption{Test error Lipschitz regularized network with parameter 0.005, $2\%$ noise data, second component.}
\end{subfigure}
\caption{Test error comparison, $2\%$ noise in the data.}
\label{fig:Err2NPend1}
\end{figure}
\end{document}